%% file: eccv2020submission.tex
\documentclass[runningheads]{llncs}
\usepackage{graphicx}
\usepackage{tikz}
\usepackage{comment}
\usepackage{amsmath,amssymb} % define this before the line numbering.
\usepackage{color}
\usepackage{booktabs}
\usepackage{makecell}
\begin{document}
% \linenumbers
\pagestyle{headings}
\mainmatter
\def\ECCVSubNumber{2136}  % Insert your submission number here

\title{3D Human Shape Reconstruction from a Polarization Image}
%\title{Detailed Human Shape and Pose Estimation from a Single Polarization Image}

\begin{comment}
\titlerunning{ECCV-20 submission ID \ECCVSubNumber} 
\authorrunning{ECCV-20 submission ID \ECCVSubNumber} 
\author{Anonymous ECCV submission}
\institute{Paper ID \ECCVSubNumber}
\end{comment}
%******************

%\begin{comment}
\titlerunning{3D Human Shape Reconstruction from a Polarization Image}
\author{Shihao Zou\inst{1} \and
Xinxin Zuo\inst{1} \and
Yiming Qian\inst{2} \and
Sen Wang\inst{1} \and
Chi Xu\inst{3} \and
Minglun Gong\inst{4} \and
Li Cheng\inst{1}}
\authorrunning{S. Zou et al.}

\institute{University of Alberta \and Simon Fraser University \and China University Of Geosciences \and University of Guelph\\
\email{\{szou2,xzuo,sen9,lcheng5\}@ualberta.ca,yimingq@sfu.ca,\\xuchi@cug.edu.cn,minglun@uoguelph.ca}}
%\end{comment}

\maketitle
\begin{abstract}
This paper tackles the problem of estimating 3D body shape of clothed humans from single polarized 2D images, i.e. polarization images. Polarization images are known to be able to capture polarized reflected lights that preserve rich geometric cues of an object, which has motivated its recent applications in reconstructing surface normal of the objects of interest. Inspired by the recent advances in human shape estimation from single color images, in this paper, we attempt at estimating human body shapes by leveraging the geometric cues from single polarization images. 
%This has led to the curation of Polarization Human Shape and Pose Dataset (PHSPD), our home-grown polarization image dataset of various human shapes and poses\footnote{Details of our dataset can be found in \url{https://jimmyzou.github.io/publication/2020-PHSPDataset}.}. 
A dedicated two-stage deep learning approach, SfP, is proposed: given a polarization image, stage one aims at inferring the fined-detailed body surface normal; % that retains individuals' fine surface shapes including clothing details; 
stage two gears to reconstruct the 3D body shape of clothing details. % based on a rough shape model (SMPL). 
Empirical evaluations on a synthetic dataset (SURREAL) as well as a real-world dataset (PHSPD) demonstrate the qualitative and quantitative performance of our approach in estimating human poses and shapes. 
This indicates polarization camera is a promising alternative to the more conventional color or depth imaging for human shape estimation. 
Further, normal maps inferred from polarization imaging play a significant role in accurately recovering the body shapes of clothed people. 
%The intuition is that the polarization camera (images) can capture the reflected light (polarized) that conveys rich geometric cues of an object. We propose a learning method to estimate normal map based on these physical clues from polarization images as priors. Then we predict the SMPL human shape as the base shape and deform it with the predicted surface normal to obtain the human shape with geometric details. We acquire a dataset, Polarization Human Shape and Pose Dataset (PHSPD)\footnote{Our dataset will be public available for research purpose upon acceptance.}, which has 12 subjects, 18 types of actions and 287K frames in total. We propose an economical yet effective way to annotate the SMPL shape and the 3D pose. 
%The experiment on our real dataset PHSPD and the synthetic dataset SURREAL shows that we are able to obtain more accurate normal maps from the polarization image than the color images, and also the normal maps helps to recover detailed human shapes.

\end{abstract}

\section{Introduction}
% \xinxin{should we change a single polarization image into a single shot...?}
% human shape and its detailed shape
Compared to the task of color-image based pose estimation~\cite{park20163d,li2015maximum,tekin2016structured,tome2017lifting,martinez2017simple,zhao2017simple,moreno20173d,nie2017monocular,zhou2017towards,wang2018drpose3d,yang20183d,fang2018learning,pavlakos2018ordinal,sun2018integral,liu2019feature,sharma2019monocular,habibie2019wild,wandt2019repnet,li2019generating,wang20193d} that predicts 3D joint positions of an articulated skeleton, human shapes provide much richer information of a human body in 3D and are visually more appealing.
%Estimating 3D human shapes %with geometric details 
%from single RGB images has been 
It, on the other hand, remains a challenging problem, %mainly due to the lack of 3D related information. 
partly owing to the relative high-dimensional space of human body shapes.
%However, because of the high-dimension of human shape, it is challenging to estimate it from only a single image. 
% \wu{Change the following two sentences to: SCAPE \cite{anguelov2005scape} and SMPL \cite{loper2015smpl} learn statistical models of low-dimensional representations of human shape from a large number of scanned body shapes. Hinging on SCAPE and SMPL, end-to-end deep learning methods to estimate human shape from a single image have been proposed.} 
The issue is somewhat alleviated by the emerging low-dimensional modelling of human shape, such as SCAPE~\cite{anguelov2005scape} and SMPL~\cite{loper2015smpl}, statistical models that are learned from large sets of carefully scanned 3D body shapes. Based on these low-dimensional human shape representations, a number of end-to-end deep learning methods~\cite{balan2007detailed,dibra2017human,dibra2016hs,bogo2016keep,lassner2017unite,kanazawa2018end,varol2017learning,pavlakos2018learning,omran2018neural,xu2019denserac,kolotouros2019learning,zanfir2018monocular,sun2019human,kanazawa2019learning,arnab2019exploiting} are subsequently developed to estimate human shapes directly from color images. 
The predicted human shapes, however, are usually naked and lacking in surface details, since e.g. SMPL model is learned from naked human body scans. 

Volume-based techniques~\cite{varol2018bodynet,zheng2019deephuman} are widely used in capturing surface details of a clothed human body from a single image. Due to finite computational resource, the estimated human shapes from these methods are usually of low resolution. Saito et al.~\cite{saito2019pifu} consider to remedy this by predicting a pixel-aligned implicit surface function that captures more detailed body surface. It however relies on a large training set of detailed 3D human bodies, and the method is still unable to handle complex poses. In the meantime, the methods of~\cite{tang2019neural} and \cite{zhu2019detailed} aim to exploit additional geometric cues arising from color image inputs; \cite{tang2019neural} instead focuses on predicting fine depth maps, and \cite{zhu2019detailed} takes on the shading aspect. Unfortunately, accurate and reliable prediction of these geometric cues from a color image is yet another challenging issue - it remains unclear how much one can leverage from such cues. 
%for complex scenes -- and this may severely hinder the applicability of theses efforts. 
%however, the prediction of an accurate base depth can be difficult, which makes the later refinement process trivial. 
%In addition, the authors in \cite{zhu2019detailed} proposed to exploit the shading information in a color image to recover the surface geometric details. However, the accuracy of the ground-truth depth maps used to train the shading network is not guaranteed and it is rather difficult to extract shading information from images as affected by the textures of the object as well as complex environmental lighting. 
%Therefore, the predicted surface details can be unreliable or even incorrect. 
Motivated by these efforts and their limitations, we consider in this paper to work with a new 2D imaging modality, polarization camera, that is known at better preserving fine-scale geometric properties of 3D objects, including human shapes. 
%To overcome the issues of previous works, in this paper, we propose to estimate the detailed human shape and pose from a single polarization image. 
The intuition comes from basic physics principle: when a light ray reflects off an object, it is polarized and conveys ample geometric cues concerning local surface details of the object, usually represented as surface normal~\cite{yang2018polarimetric,ba2019physics}. It may be found to note some biological species are even able to directly perceive light polarization~\cite{wehner2006significance,daly2016dynamic}, which significantly facilitates their 3D sensing. Empirically, our experiments support that the surface normal maps obtained out of the input 2D polarization images could play an instrumental role in producing accurate and reliable 3D 
clothed human shapes.

\begin{figure}[htb]
    \centering
    \includegraphics[width=\columnwidth]{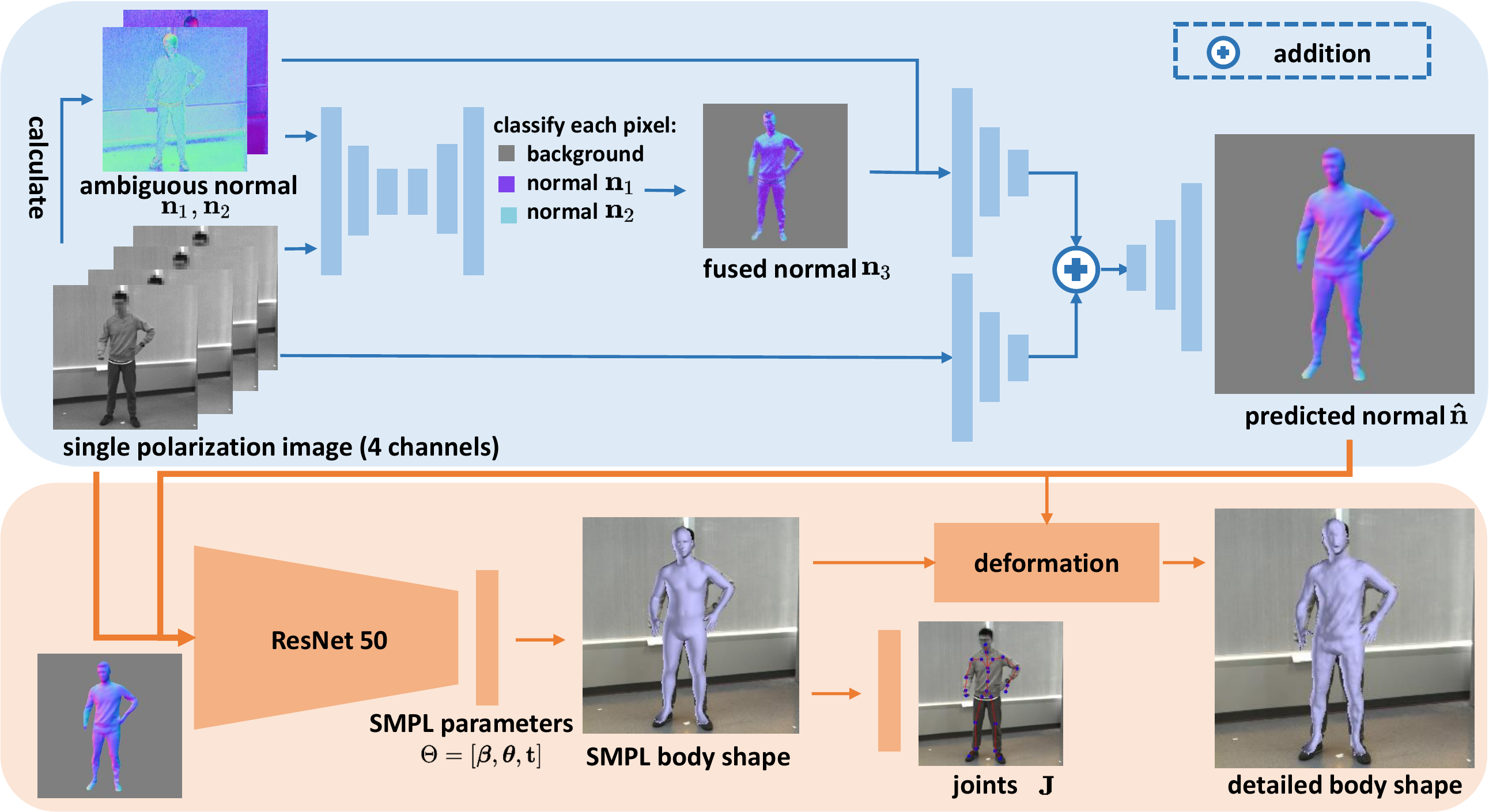}
    \caption{Given a single polarization image, a two-stage process is executed in our approach. (1) Stage 1, in blue, estimates the surface normal from the polarization image based on the physical assumption that reflected light from an object is polarized. After calculating the two ambiguous normal maps, ($\mathbf{n}_1$, $\mathbf{n}_2$), as physical priors from the polarization image (see Sec.~\ref{sec:normal-estimation} for details), image pixels are classified as belonging to either of the two normals or a background, thus obtaining the fused normal $\mathbf{n}_3$. Unfortunately, this normal is often noisy, thud a further step is carried out in regressing a final accurate surface normal $\hat{\mathbf{n}}$, by integrating these physical normal maps and the raw polarization image.
    %as an extra \emph{improved} physical prior. 
    %The final surface normal is predicted based on these physical priors and the raw image. 
    (2) Stage 2, in orange, concatenates the polarization image and the surface normal as the input to estimate clothed body shape in two steps. The first step focuses on estimating the parameters of SMPL, a rough \& naked shape model parameterized by $\Theta$; the pose (3D joint positions) $\mathbf{J}$ is directly obtained as a by-product of the rigged shape model. The next step deforms the SMPL shape guided by the final surface normal of stage 1, to reconstruct the refined 3D human shape with clothing details.}
    \label{fig:overview}
\end{figure}

As shown in Fig.~\ref{fig:overview}, our approach, also called SfP, contains two stages. Stage 1 concentrates on predicting accurate surface normal maps from single polarization images\footnote{In this paper, an polarization image has four channels with each channel corresponding to a specific polarizer degree of (0, 45, 90 and 135).} by exploiting the associated physics laws as priors. It is then fed into stage 2 in reconstructing the final clothed human shape. 

Unlike existing efforts in normal map prediction~\cite{tang2019neural,zhu2019detailed}, our approach predict normal maps by explicitly incorporating the underlying physical laws of polarization imaging, which results in more \emph{reliable} performance. To achieve this, there are two main challenges we need to overcome, namely $\pi$-ambiguity of the azimuth angle and the possibly large noise in practical applications. To this end we introduce two ambiguous normal maps $\mathbf{n}_1$ and $\mathbf{n}_2$ (Sec. \ref{sec:normal-estimation}) as a physical prior, based on the assumption that the light reflected by human clothing is mostly diffused. Different from~\cite{ba2019physics}, each pixel is then classified into one of the three types: the two ambiguous normal maps and background. This is followed by a refinement step to deliver the final surface normal prediction of $\hat{\mathbf{n}}$, that accounts for the possibly-noisy fused normal map output owing to environmental noise and the digital quantization of the polarization camera. 
%the fused normal map from the classification result is noisy. Employing the fused normal as an extra \emph{improved} physical prior, we train a denoising network to refine the normal map.
% Then because of the environmental noise and the digital quantization of the camera, the fused normal map from the classification result are noisy. 
% \wu{Due to environmental noise and the digital quantization of the camera, the fused normal map from the classification result is noisy. Employing the fused normal as an extra \emph{improved} physical prior, we train a denoising network to predict the normal map.} 
%With the fused normal as an extra \emph{improved} physical prior, we train a denoising network to predict the normal map.
%
Based on the raw polarization image and output of stage 1, stage 2 concerns the estimation of clothed human shape. It starts from predicting a coarse SMPL shape model, which is then deformed by leveraging the geometric details from surface normal, our stage 1 output, to form the final human shape. Empirically our two-stage pipeline is shown to be capable of accurately reconstructing human shapes, while retaining clothing details such as cloth wrinkles.
%The predicted normal map which implies 3D information of the surface provides useful clues for shape estimation than a single polarization or color image and better performance has been achieved when the normal map is also involved as a part of the input. Finally the detailed human body shape is reconstructed by deforming the base SMPL shape towards the predicted normal map. As demonstrated in the experiments, we are able to obtain real geometric details of the human body such as the wrinkles of the cloth from polarization images. 
%\xinxin{Do we have to repeat this here?} We argue that compared with the direct depth regression in \cite{tang2019neural}, it is more robust by using the SMPL shape as the base shape, especially for complex poses. On the other hand, unlike \cite{zhu2019detailed} that exploits shading information in the image which could be sensitive to environment lighting and surface texture and finally results in fake or hallucinate surface details, we rely on the normal map estimated from polarization images to recover real geometry details in our deformation stage.

To summarize, there are two main contributions in this work. (1) A new problem of inferring high-resolution 3D human shapes from a single polarization image is proposed and investigated. 
This lead us to curate a dedicated Polarization Human Shape and Pose Dataset (PHSPD). %, which consists of 12 subjects, 18 types of actions and 287K frames in total, with each frame including 7 synchronized images: one polarization image, three-view color and depth images. 
%To the best our knowledge, our work is the first one that investigates the new question of whether the geometric cues from polarization camera could be leveraged to estimate \emph{detailed} human shapes and pose. 
(2) A dedicated deep learning approach, SfP, is proposed\footnote{Our project website is https://jimmyzou.github.io/publication/2020-polarization-clothed-human-shape}, where the detail-preserving surface normal maps are obtained following the physical laws, and are shown to significantly improve the reconstruction performance of clothed human shapes. Empirical evaluations on a synthetic SURREAL dataset as well as a real-world dataset demonstrate the applicability of our approach. Our work provide sound evidence in engaging 2D polarization camera to estimate 3D human poses and shapes, a viable alternative to conventional 2D color or 3D depth cameras.
%to annotate SMPL shape parameters and 3D pose (joint positions), which does not require subjects to wear many sensors and special tight clothes.
%Empirical evaluations on a synthetic SURREAL dataset as well as our PHSPD dataset demonstrate that the normal maps estimated from polarization images play an instrumental role in producing accurate and reliable reconstruction of 3D human shapes, which shows the advantages of polarization images to estimate detailed human shape and pose.

\section{Related Work}
\paragraph{Shape from polarization (SfP)} focuses on the inference of shape (normally represented as \emph{surface normal}) from the polarimetric information in the multiple channels of a polarization image, captured under linear polarizers with different angles. The main issue of SfP is angle ambiguity. Previous methods are mainly physics-based that rely on other additional information or assumptions to elucidate the possible ambiguities, such as smooth object surfaces \cite{atkinson2006recovery}, coarse depth map \cite{kadambi2017depth,yang2018polarimetric} and multi-view geometric constraint \cite{chen2018polarimetric,cui2017polarimetric}. The recent work of~\cite{ba2019physics} proposes to blend physical priors (ambiguous normal maps) with deep learning in uncovering the normal map. Using physical priors as part of the input, deep learning model can then be trained to account for the ambiguity and be noise-resilient. We improve upon~\cite{ba2019physics} by classifying ambiguous normal and background for each pixel, and regressing the normal given the ambiguous and classified physical priors.

\paragraph{3D human pose estimation from single images} has been extensively investigated in the past five years, centering around color or depth imaging. Many of the studies~\cite{zhou2016sparseness,akhter2015pose,wang2014robust,ramakrishna2012reconstructing,zhou2018monocap,zhou2016sparse,chen20173d} utilize dictionary-based learning strategies. More recent efforts aim to directly regress 3D pose using deep learning techniques, including CNNs~\cite{park20163d,li2015maximum,tekin2016structured} and Graph CNNs~\cite{ci2019optimizing,cai2019exploiting}. In prticulr, several recent efforts~\cite{tome2017lifting,martinez2017simple,zhao2017simple,moreno20173d,nie2017monocular,zhou2017towards,wang2018drpose3d,yang20183d,fang2018learning,fang2018learning,pavlakos2018ordinal,sun2018integral,liu2019feature,sharma2019monocular,habibie2019wild,wandt2019repnet,li2019generating,wang20193d} look into a common framework of estimating 2D pose (either 2D joint positions or heatmap), which is then lifted to 3D. Ideas from self-supervised learning \cite{wang20193d,habibie2019wild} and adversarial learning \cite{yang20183d,wandt2019repnet} also gain attentions in e.g. predicting 3D pose under additional constraints imposed from re-projection or adversarial losses. %Our method departs from these previous works in that we are estimating human shape and pose from polarization images and we argue that the reliable normal map obtained from polarization images as an informative prior can help to achieve better human shape and pose estimation.

\paragraph{Human shape estimation from single images} has drawn growing attentions recently, thanks to development of human shape models of SCAPE and SMPL \cite{anguelov2005scape,loper2015smpl}. These two statistical models learn low-dimensional representations of human shape from large corpus of human body scans. Together with deep learning techniques, it has since been feasible to estimate human body shapes from single color or depth images. Earlier activities focus more on optimizing the SCAPE or SMPL model parameters toward better fitting to various dedicated visual or internal representations, such as foreground silhouette~\cite{balan2007detailed,dibra2017human,dibra2016hs} and pose~\cite{bogo2016keep,lassner2017unite}. Deep learning based approaches are more commonplace in recent efforts~\cite{kanazawa2018end,varol2017learning,pavlakos2018learning,omran2018neural}, which typically learn to predict the SMPL parameters by incorporating the constrains from 2/3D pose, silhouette, as well as adversarial learning losses. \cite{xu2019denserac} takes the body pixel-to-surface correspondence map as proxy representation and then performs estimation of parameterized human pose and shape. In \cite{kolotouros2019learning}, optimization-based methods \cite{bogo2016keep} and regression-based methods \cite{kanazawa2018end} are combined to form a self-improved fitting loop. point cloud is considered as input in~\cite{jiang2019skeleton} to regress SMPL parameters. Instead of single color images, our work is based on single polarization image; rather than inferring coarse human body shape, we aim to recover high-res human shapes. 
%We further deform the shape with the surface normal predicted from a single polarization image to produce detailed human shape.

% There are also some works that introduce temporal constrains to get more smooth human shape estimation. \cite{zanfir2018monocular,sun2019human,kanazawa2019learning,arnab2019exploiting}

As for the estimation of clothed human shape, volume-based methods~\cite{varol2018bodynet,zheng2019deephuman,saito2019pifu} are proposed to reconstruct textured body shapes. 
%Due to the limitation of computational resource, 
they unfortunately suffer from the low resolution issue of volumetric representation. Our work is closely related to \cite{zhu2019detailed}, which combines the robustness of parametric model and the flexibility of free-form 3D deformation in a hierarchical manner.
% , utilizing the constraints from body joints, silhouettes, and per-pixel shading information to recover the human shape with details. 
The major difference is, the clothing details of our work are provided by the reliable normal map estimated from the polarization image, whereas the network in \cite{zhu2019detailed} deforms depth image by employing the shading information trained on additional data, %an extra small set of data 
that are inherently unreliable due to the lack of ground-truth information of surface normal, albedo and environmental lighting. 
Our work is also related to \cite{tang2019neural} which recovers detailed human shape from a color image, by iteratively incorporating both rough depth map and estimated surface normal for improved surface details. %The drawback of \cite{tang2019neural} is that the poses in their dataset are simple. Therefore, in the case of complex poses, the prediction of base depth may be inaccurate, leading to the failure of deformation. On the contrary, we obtain the base depth by rendering predicted SMPL shape, which is robust to complex poses.

\section{The Proposed SfP Approach}
There are two main stages in our approach: (1) estimate surface normal from a single polarization image; (2) estimate human pose and shape from the estimated surface normal and the raw polarization image, followed by body shape refinement from the estimated surface normal.
% overview

\subsection{Surface Normal Estimation}
\label{sec:normal-estimation}
The reflected light from a surface mainly includes three components~\cite{cui2017polarimetric}, the polarized specular reflection, the polarized diffuse reflection, and the unpolarized diffuse reflection. 
A polarization camera has an array of linear polarizer mounted right on top of the CMOS imager, similar to the RGB Bayer filters. During the imaging process of a polarization camera, a pixel intensity typically varies sinusoidally with the angle of the polarizer~\cite{yang2018polarimetric}.
%Normally, a single shot of the polarization camera can capture the illumination intensity under four different polarizer angles (with each channel corresponding to one polarizer angle). 
%
In this work, we assume that the light reflected off human clothes is dominated by polarized diffuse reflection and unpolarized diffuse reflection. For a specific polarizer angle $\phi_{\mathrm{pol}}$, the illumination intensity at a pixel with dominant diffuse reflection is
\begin{equation}
    \label{eq:polar}
    \mathrm{I}(\phi_{\mathrm{pol}})=\frac{\mathrm{I}_{\max} + \mathrm{I}_{\min}}{2} + \frac{\mathrm{I}_{\max} - \mathrm{I}_{\min}}{2} \cos (2(\phi_{\mathrm{pol}}-\varphi)).
\end{equation}
Here $\varphi$ is the azimuth angle of surface normal, $\mathrm{I}_{\max}$ and $\mathrm{I}_{\min}$ are the upper and lower bounds of the illumination intensity. $\mathrm{I}_{\max}$ and $\mathrm{I}_{\min}$ are mainly determined by the unpolarized diffuse reflection, and the sinusoidal variation is mainly determined by the polarized diffuse reflection. Note that there is $\pi$-ambiguity in the azimuth angle $\varphi$ in Eq.~(\ref{eq:polar}), which means that $\varphi$ and $\pi+\varphi$ will result in the same illumination intensity of the pixel. As for the zenith angle $\theta$, it is related to the degree of polarization $\rho$, where
\begin{equation}
    \label{eq:degree-of-polarization}
    \rho = \frac{\mathrm{I}_{\max} - \mathrm{I}_{\min}}{\mathrm{I}_{\max} + \mathrm{I}_{\min}}.
\end{equation}
According to \cite{atkinson2006recovery}, when diffuse reflection dominates, the degree of polarization $\rho$ is a function of the zenith angle $\theta$ and the refractive index $n$,
\begin{equation}
    \label{eq:zenith-angle}
    \rho = \frac{(n-\frac{1}{n})^2\sin^2\theta}{2+2n^2-(n+\frac{1}{n})^2\sin^2\theta+4\cos\theta\sqrt{n^2-\sin^2\theta}}.
\end{equation}
In this paper, we assume the refractive index $n=1.5$ since the material of human clothes is mainly cotton or nylon. With $n$ known, the solution of $\theta$ in Eq.~(\ref{eq:zenith-angle}) is a close-form expression of $n$ and $\rho$.

Taking into account the $\pi$-ambiguity of $\varphi$, we have two possible solutions to the surface normal for each pixel, that form the physical priors. We propose to train a network to classify each pixel into three categories: background, ambiguous normal $\mathbf{n}_1(\varphi, \theta)$  and ambiguous normal $\mathbf{n}_2(\pi+\varphi, \theta)$ with probability $p_0$, $p_1$, and $p_2$ respectively. Then we have the fused normal as follows,
\begin{equation}
    \label{eq:fuse-noisy-normal}
    \mathbf{n}_3 = (1-p_0) \cdot \frac{p_1 \mathbf{n}_1 + p_2 \mathbf{n}_2}{\|p_1 \mathbf{n}_1 + p_2 \mathbf{n}_2\|_2},
\end{equation}
where $(1-p_0)$ is a soft mask of the foreground human body. Unfortunately, due to the environmental noise and the digital quantization of camera in real-world applications, the fused normal map $\mathbf{n}_3$ is noisy and non-smooth. Thus taking the fused noisy normal as an \emph{improved} physical prior, a denoising network is further trained to take both the polarization image and the physical priors $(\mathbf{n}_1, \mathbf{n}_2, \mathbf{n}_3)$ as input, and to produce a smoothed normal $\mathbf{\hat n}$. The loss function for normal estimation consists of the cross entropy (CE) loss of classification and the L1 loss of the cosine similarity,
\begin{align}
    \label{eq:normal-loss}
    L_{n} &= \frac{1}{HW}\sum_{i=1}^{H}\sum_{j=1}^{W} \left[ \lambda_{c}\text{CE}(y^{i, j},p^{i,j})+  \lambda_{n}(1-\langle\mathbf{\hat n}^{i, j}, \mathbf{n}^{i, j}\rangle)\right],
\end{align}
where $\lambda_{c}$ and $\lambda_{n}$ are the weights of each loss, $y^{i, j}$ is the label indicating which category the pixel $(i,j)$ belongs to, and $\langle\mathbf{\hat n}^{i, j}, \mathbf{n}^{i, j}\rangle$ denotes the cosine similarity between the predicted and target normal vectors of pixel $(i,j)$. Note that the category label $y^{i, j}$ is created by discriminating whether the pixel is background or which ambiguous normal has higher cosine similarity with the target normal. $\lambda_c$ and $\lambda_n$ is 2 and 1 respectively in our experiment.

\subsection{Human Pose and Shape Estimation}
To start with, the SMPL~\cite{loper2015smpl} representation is used for describing 3D human shapes, which is a differentiable function $\mathcal{M}(\boldsymbol{\beta}, \boldsymbol{\theta})\in\mathbb{R}^{6,890\times3}$ that outputs a triangular mesh with 6,890 vertices given 82 parameters $[\boldsymbol{\beta}, \boldsymbol{\theta}]$. The shape parameter $\boldsymbol{\beta}\in\mathbb{R}^{10}$ is the linear coefficients of a PCA shape space that mainly determines individual body features such height, weight and body proportions. The PCA shape space is learned from a large dataset of body scans \cite{loper2015smpl}. $\boldsymbol{\theta}\in\mathbb{R}^{72}$ is the pose parameter that mainly describes the articulated pose, consisting of one global rotation of the body and the relative rotations of 23 joints in axis-angle representation. Finally, our clothed body shape is produced by first applying shape-dependent and pose-dependent deformations to the template pose, then using forward-kinematics to articulate the body shape back to its current pose, and deforming the surface mesh by linear blend skinning.  $\mathbf{J}\in\mathbb{R}^{24\times3}$ are the 3D joint positions that can be obtained by linear regression from the output mesh vertices. 

In addition to the SMPL parameters, we also need to predict the global translation $\mathbf{t}\in\mathbb{R}^3$. Thus for the task of human pose and shape estimation, 
%we concatenate \emph{one polarization image and also the predicted surface normal together as the input} and 
the output vector is of 85-dimension, $\hat \Theta=[\boldsymbol{\hat\beta}, \boldsymbol{\hat\theta}, \mathbf{\hat t}]$. 
%The empirical results show that the normal map plays an instrumental role to estimate better human shapes and poses. 
Given $\hat \Theta$, we can also obtain the predicted 3D joint positions $\mathbf{\hat J}$. To this end, the loss function is defined as
\begin{align}
    \label{eq:shape-loss}
    L_{s} &= \lambda_{\beta}\|\boldsymbol{\beta} - \boldsymbol{\hat \beta}\|^2_2 + \lambda_{\theta}\|\boldsymbol{\theta} - \boldsymbol{\hat \theta}\|^2_2 + \lambda_{t}\|\mathbf{t} - \mathbf{\hat t}\|^2_2 + \lambda_J \|\mathbf{J} - \mathbf{\hat J}\|^2_2,
\end{align}
where $\lambda_{\beta}$, $\lambda_{\theta}$, $\lambda_{t}$ and $\lambda_{J}$ are weights of each component in the loss function, which are fixed to 0.2, 0.5, 100, and 3, respectively. 

The reconstructed SMPL human shape thus far is naked 3D shape and lacking fine surface details. Our goal is to refine this intermediate naked shape under the guidance of our smoothed surface normal estimate. It is carried out as follows. The SMPL body shape is rendered on the image plane to form a base depth map. The technique in \cite{nehab2005efficiently} is then engaged here to obtain an optimized depth map from the predicted surface normal and the base depth map. It is carried out under three constraints: first, the predicted normal should be perpendicular to the local tangent of the optimized depth surface; second, the optimized depth should be close to the base depth; Third, a smoothness constraint is enforced on nearby pixels of the optimized depth map. This depth map is obtained as a solution of a linear least-squares system. Weights of the normal term, the depth data term, and the smoothness term are empirically set to 1.0, 0.06, and 0.55, respectively. 
Finally, our clothed body shape is produced by upsampling \& deforming the SMPL mesh according to the Laplacian of the optimized depth map.
%Then we attach the integrated depth map onto the naked human mesh by deforming the human mesh with Laplacian deformation controlled by the correspondences from the vertices on the human mesh to the integrated depth map. We upsample the human mesh before the deformation so as to represent the surface details.

\subsection{Polarization Human Pose and Shape Dataset}
To facilitate empirical evaluation of our approach in real-world scenarios, a home-grown dataset is curated, referred as Polarization Human Shape and Pose Dataset, or PHSPD. 
%In what follows, a short deception of the PHSPD dataset is provided for the sake of self-completeness. 
A complete description of this PHSPD dataset is provided in~\cite{zou2020polarization}. 
%Since there is no such dataset that can be used to estimate human pose and/or shape from polarization images, we acquire such a dataset~\cite{zou2020polarization}, referred as Polarization Human Shape Dataset (PHSPD). 
In data Requisition stage, a system of four soft-synchronized cameras are engaged, consisting of a polarization camera and three Kinects V2, with each Kinect v2 having a depth and a color cameras. 12 subjects are recruited in data collection, where 9 are male and 3 are female. Each subject performs 3 different groups of actions (out of 18 different action types) 4 times, plus an addition period of free-form motion at the end of the session. Thus for each subject, there are 13 short videos (of around 1,800 frames per video with 10-15 FPS); the total number of frames for each subject amounts to 22K. Overall, our dataset consists of 287K frames, each frame here contains a synchronized set of images - one polarization image, three color and three depth images.

\begin{figure}[htb]
    \centering
    \includegraphics[width=\columnwidth]{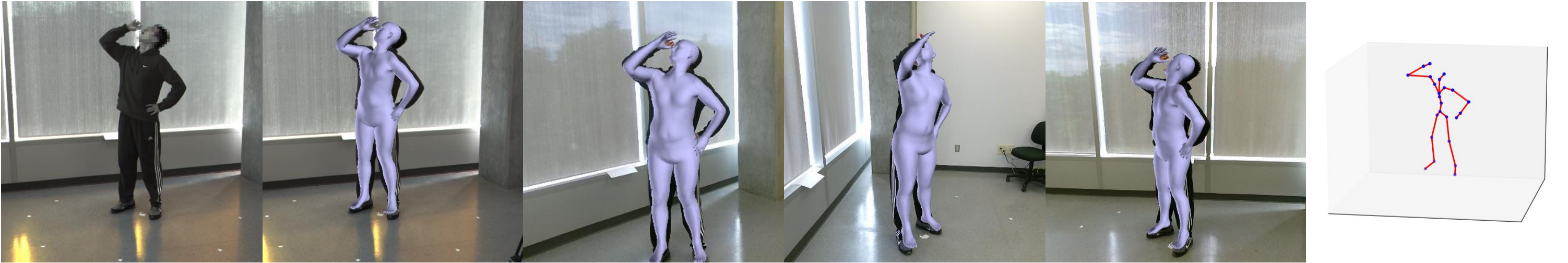}
    \includegraphics[width=\columnwidth]{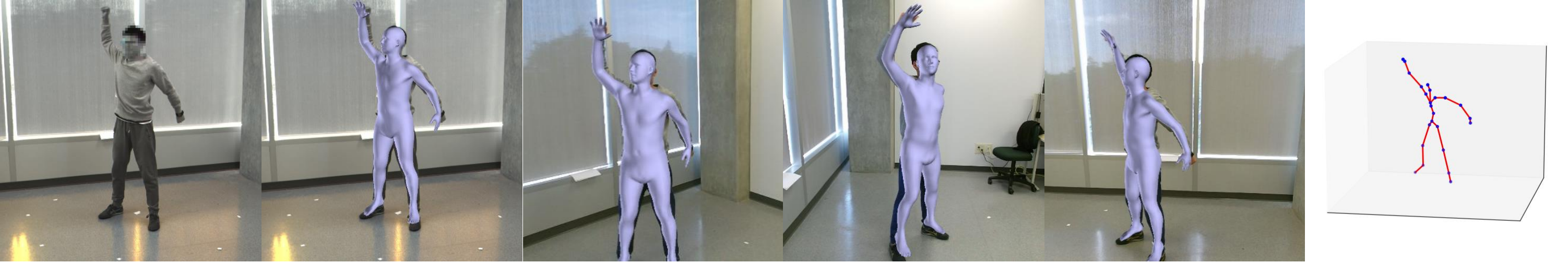}
    \includegraphics[width=\columnwidth]{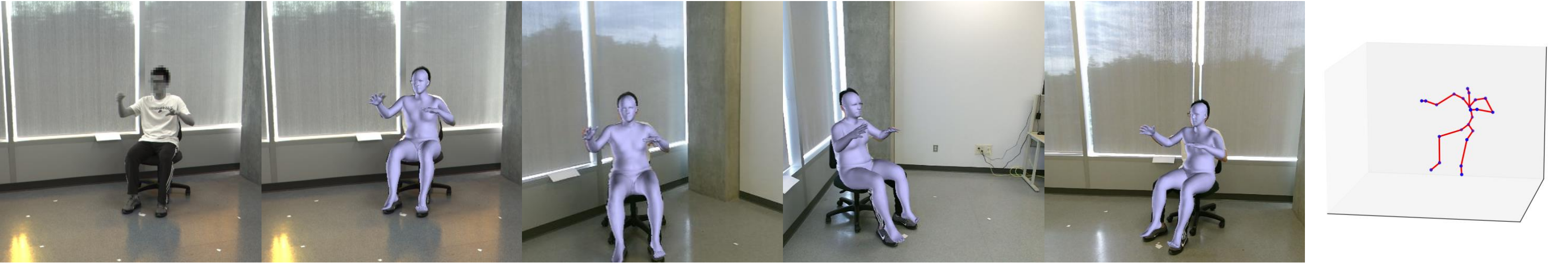}
    \caption{Exemplar 3D poses and SMPL shapes in the real-world PHSPD dataset. We render the SMPL shape on four images (one polarization image and three-view color images) and we also show the pose in 3D space.}
    \label{fig:multiview-demo}
\end{figure}

The SMPL shape parameters and the 3D joint positions of a body shape are obtained from the image collection of current frame as follows. 
%our first step is to obtain the initial guess of the 3D joint positions. 
For each frame, its initial 3D pose estimation is obtained by integrating the Kinect readouts as well as the corresponding 2D joint estimation from OpenPose~\cite{openpose} across the depth and color sensors. 
%After converting the accurate 3D joint estimations of three Kinects to the polarization camera coordinate, the initial guess is obtained by taking the average of these accurate estimations. We discard frames with more than two missing joints (inaccurate in all the three views). 
Then the body shape, i.e. parameters of the SMPL model, is estimated as optimal fit to the initial pose estimate~\cite{bogo2016keep}. 
The 3D point cloud of body surface acquired from three depth cameras are now utilized in our final step, resulting in the estimation of refined body shape with clothing details~\cite{zuo2020sparsefusion}, by iteratively minimizing the distance of SMPL shape vertex to its nearest point of the 3D point cloud.
%is to further optimize SMPL parameters by iteratively minimizing the distance between vertices of SMPL shape to their nearest vertices on the point cloud.
%Finally, we will have the annotated SMPL models that fits closely to the captured human surface and also the 3D joint positions. 
Exemplar clothed human shapes are shown in Fig.~\ref{fig:multiview-demo}. 
%Besides, we render the boundary of SMPL shape on the image to get the mask of foreground, and calculate the target normal from the point cloud.\cite{qi2018geonet} Although the target normal is noisy, our experiment results demonstrate our model can still learn to estimate good and smooth normal maps. 

%More details about our dataset and acquisition process can be found in supplementary material. Compared with Human3.6M \cite{human36m} that uses expensive Motion Capture system to annotate human poses, our proposed process is economical and effective. Furthermore, we do not require subjects to wear special tight clothes and a lot of sensors, which makes the acquired images restrictive and impractical.

\section{Empirical Evaluations}
%\subsection{Implementation Details}
Empirical evaluations are carried out on two major aspects. (1) For normal estimation, we report the Mean Angle Error (MAE), which measures the angle between the target and estimated normal map, $e_{\mathrm{angle}}=\arccos(\langle \mathbf{n}^{i, j}, \mathbf{\hat n}^{i, j} \rangle)$ for pixel $(i, j)$, where $\langle \cdot,\cdot \rangle$ denotes cosine similarity. (2) For human pose and shape estimation, we report the mean per joint position error (MPJPE) and the 3D surface distance error. MPJPE is defined as the average distance between predicted and annotated joints of the test samples. 
In both SURREAL and PHSPD datasets, there are 24 annotated 3D joints. We also report the MPJPE for 20 joints by removing the hand and foot joints. The 3D surface error measures the distance between the predicted mesh and the ground truth mesh, by averaged distance of vertex pairs, as follows: for each vertex of the human body mesh, its closest vertex in ground truth mesh is identified to form its vertex pair; the average distance of all such vertex pairs is then computed.

% two datasets
For the real-world PHSPD dataset, subject 4 is chosen to form the validation set (23,786 samples); the test set contains those of subjects 7, 11, and 12 (69,283 samples); the train set has everything else (186,746 samples). 
% All the images are cropped according to the background mask and resized to $256\times256$.
% \xinxin{Do you need to mention this?} 
%We assume that the camera parameters are known since the images are captured by the same camera. 
%Being a real-world dataset, we do not have a color image that exactly matches a polarization image at the same viewpoint. So a color image from a color camera having similar angle of view with the polarization camera is used. 

We also demonstrate the effectiveness of our SfP approach on SURREAL~\cite{varol2017learning}, a synthetic dataset of color images rendered from motion-captured human body shapes. Polarization images can be synthesized using color and depth images (details are in supplementary material). We choose subset "run1" and select one frame with a gap of ten frames. Finally, the train set has 245,759 samples, validation set has 14,528 samples and test set has 52,628 samples. 
% We also crop and pad the images to $256\times256$.

\subsection{Evaluation of Surface Normal Estimation}
In this task, our approach is compared with a recent work \emph{Physics}~\cite{ba2019physics}, a traditional method \emph{Linear}~\cite{smith2016linear}, and three ablation variants of our method as baselines. \emph{Ours (color image)} uses only color image for estimating the normal map. \emph{Ours (no physical priors)} does not incorporate the ambiguous normal maps as the physical priors and employs the polarization image as the only input. \emph{Ours (no fused normal)} is similar to Physics \cite{ba2019physics}, in which we use the two ambiguous normal maps as the only priors, discarding the fused normal maps.

\begin{table}[]
    \centering
    \begin{tabular}{ccc}
    \toprule
         & \quad SURREAL \quad& \quad PHSPD \quad \\
    \midrule
        Linear \cite{smith2016linear}  & 20.03 & 34.97\\
        Physics \cite{ba2019physics}  & 7.45 & 21.45\\
    \midrule
        ours (color image) & 19.49  & 25.02\\
        ours (no physical priors) & 13.89 & 24.71\\
        ours (no fused normal) & 7.43 & 21.65\\
        ours & \textbf{7.10} & \textbf{20.75}\\
        
    \bottomrule
    \end{tabular}
    \caption{Comparison of surface normal estimation evaluated in Mean Angular Error (MAE). The competing methods include \emph{Linear} \cite{smith2016linear}, \emph{Physics} \cite{ba2019physics}, \emph{ours}, and three ablation variants of our method. %Ours (color image), ours (no physical priors) and ours (no fused normal) are added as baselines to show how physical priors from polarization images contribute to improving the normal map estimation.
    }
    \label{tab:normal-estimation}
\end{table}

\begin{figure}[htb]
    \centering
    \includegraphics[width=\columnwidth]{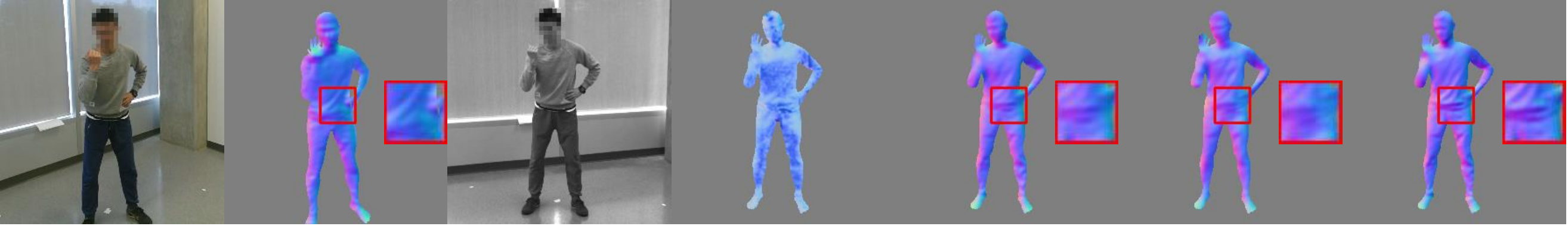}
    \includegraphics[width=\columnwidth]{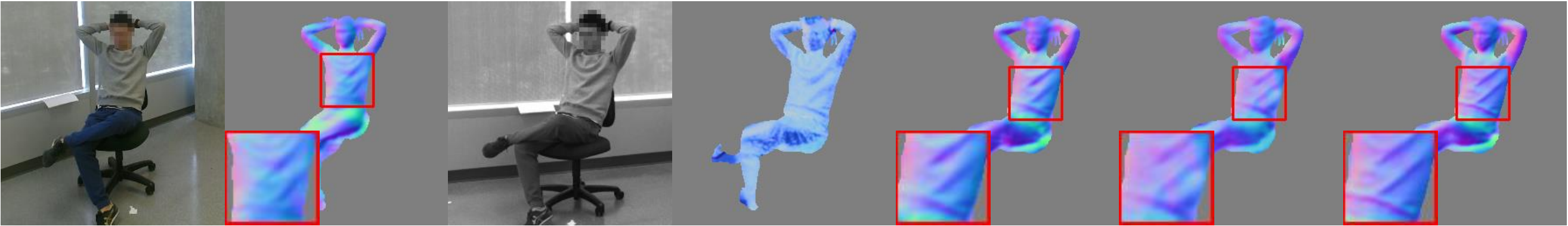}
    \includegraphics[width=\columnwidth]{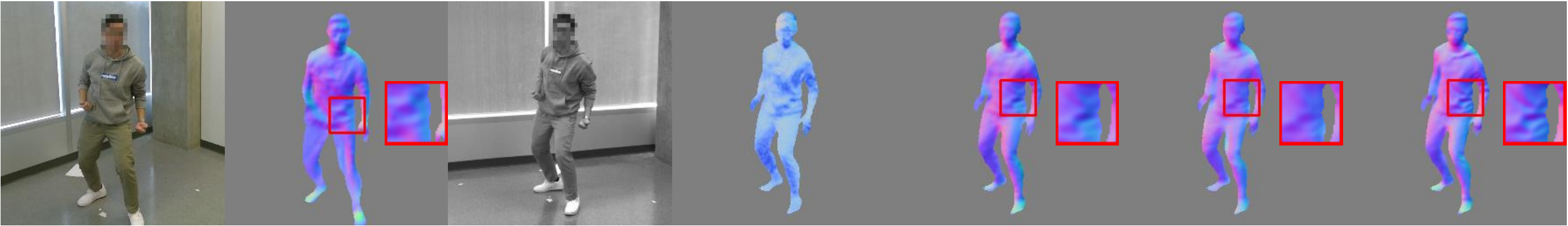}
    \includegraphics[width=\columnwidth]{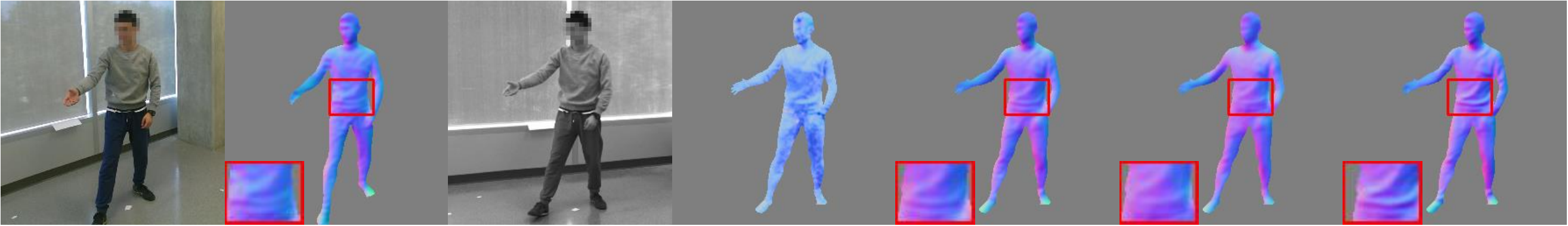}
    \includegraphics[width=\columnwidth]{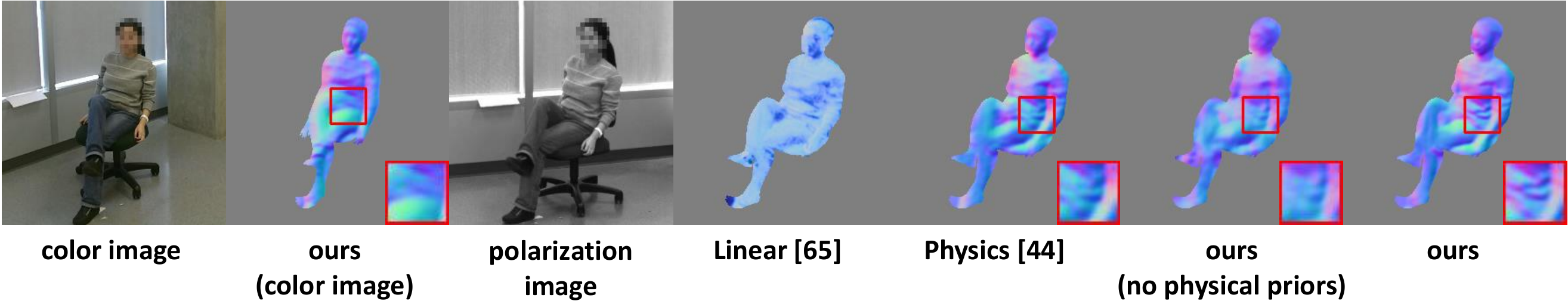}
    \caption{Exemplar results of normal map prediction by five competing methods: \cite{smith2016linear,ba2019physics}, \emph{ours (no physical priors)}, \emph{ours (color image)}, and \emph{ours}. Original color images and polarization images are shown in the first and third column with pixelated faces. 
    }
    \label{fig:normal-estimation-demo}
\end{figure}

Through both the quantitative results of Table~\ref{tab:normal-estimation} and the visual results of Fig.~\ref{fig:normal-estimation-demo}, it is observed that our method has consistently outperforms the state-of-the-art surface normal prediction methods~\cite{ba2019physics,smith2016linear} in both SURREAL and PHSPD datasets. 
The poor performance of \cite{smith2016linear} may be attributed to its unrealistic assumption of noise-free environment in the captured images. 
Let us look at the three ablation baselines of our approach: using only color images delivers relatively similar performance to that of removing physical priors when compared in PHSPD. %, which drops significantly in SURREAL. 
Intuitively, it is challenging for neural networks to encode information of ambiguous normal maps (physical priors) directly from raw polarization images. Therefore, removing the physical priors results in similar performance to that of using only color images.
%Ours (no physical priors) shows similar MAE with color images on PHSPD, but shows better results on SURREAL. Intuitively, if the ambiguous normal maps are not calculated as physical priors, it is hard for neural networks to encode such information using raw polarization images. So the result of our method (no physical priors) is close to that of color images.
% For SURREAL dataset, it is mainly because the depth images in SURREAL used to obtain target normal map are rendered from SMPL model that is naked while humans in the color images are clothed. So it is easier to predict the target normal from polarization images than color images. 
\cite{ba2019physics} and \emph{ours (no fused normal)} both utilize ambiguous normal as a physical prior, thus produce similar results. By incorporating the fused normal which discriminates the ambiguity of azimuth angle estimation, the results of our full-fledged approach surpasses those of \cite{ba2019physics}. % in terms of performance.
%We can see that adopting physical priors in our method and \cite{ba2019physics} performs better than ours (no physical priors) and ours (color image). Our method also obtains more details in the normal map estimation than \cite{ba2019physics}.

\subsection{Evaluation of Pose and Shape Estimation}
The focus of this section is qualitative and quantitative evaluations on estimating poses \& SMPL shapes, as well as our final estimation of clothed human shapes.

\begin{figure}[htb]
    \centering
    \includegraphics[width=0.49\columnwidth]{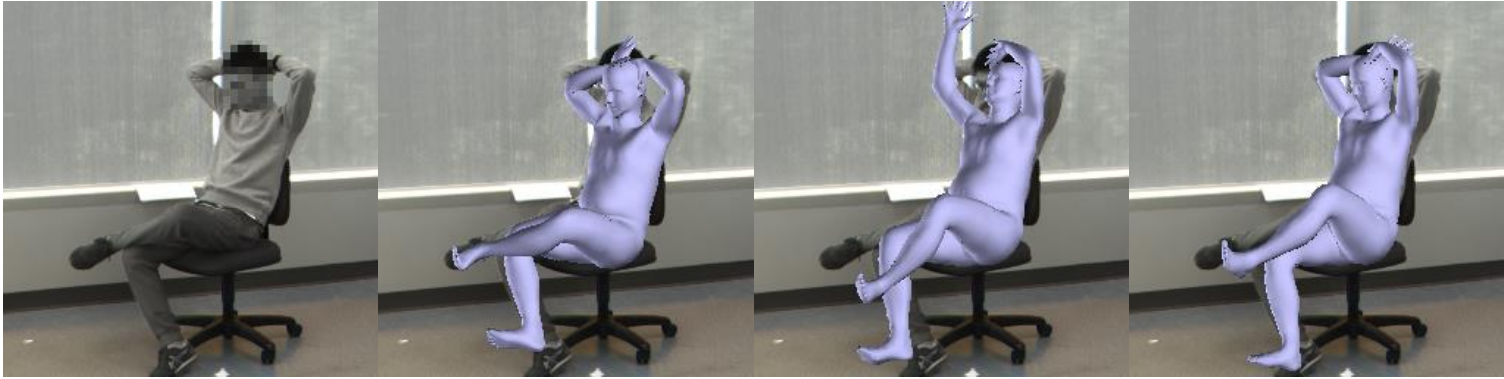}
    \includegraphics[width=0.49\columnwidth]{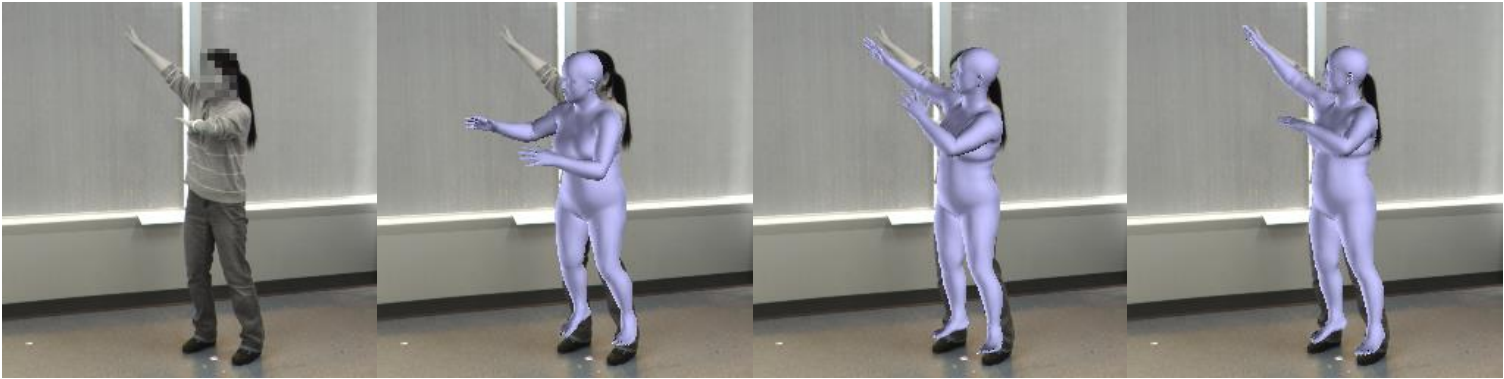}
    \includegraphics[width=0.49\columnwidth]{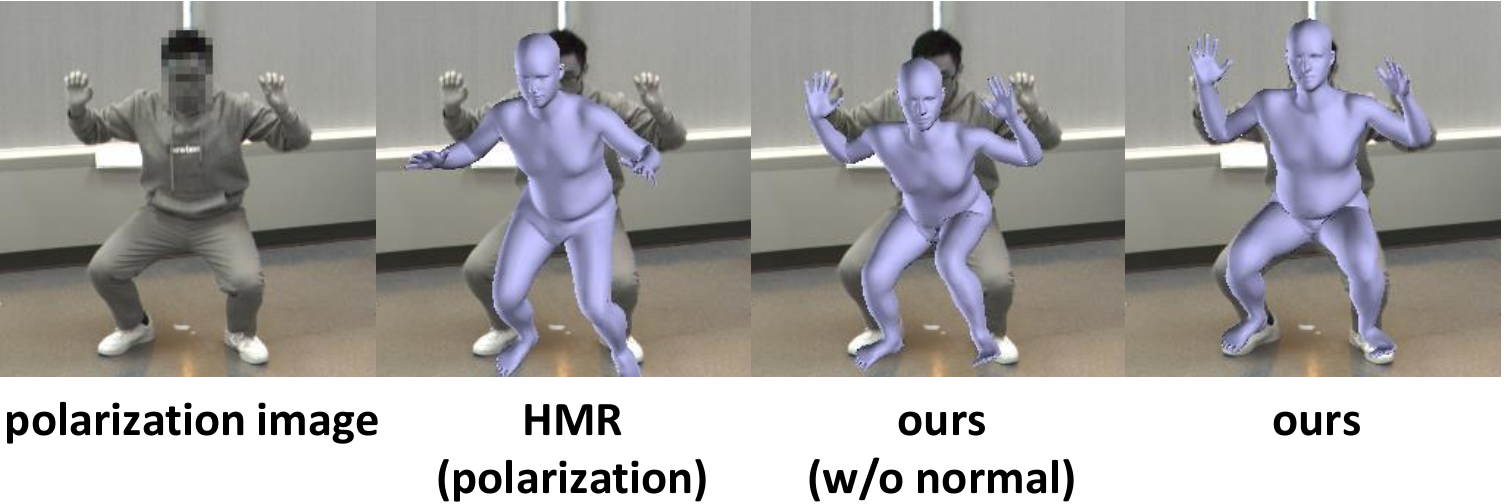}
    \includegraphics[width=0.49\columnwidth]{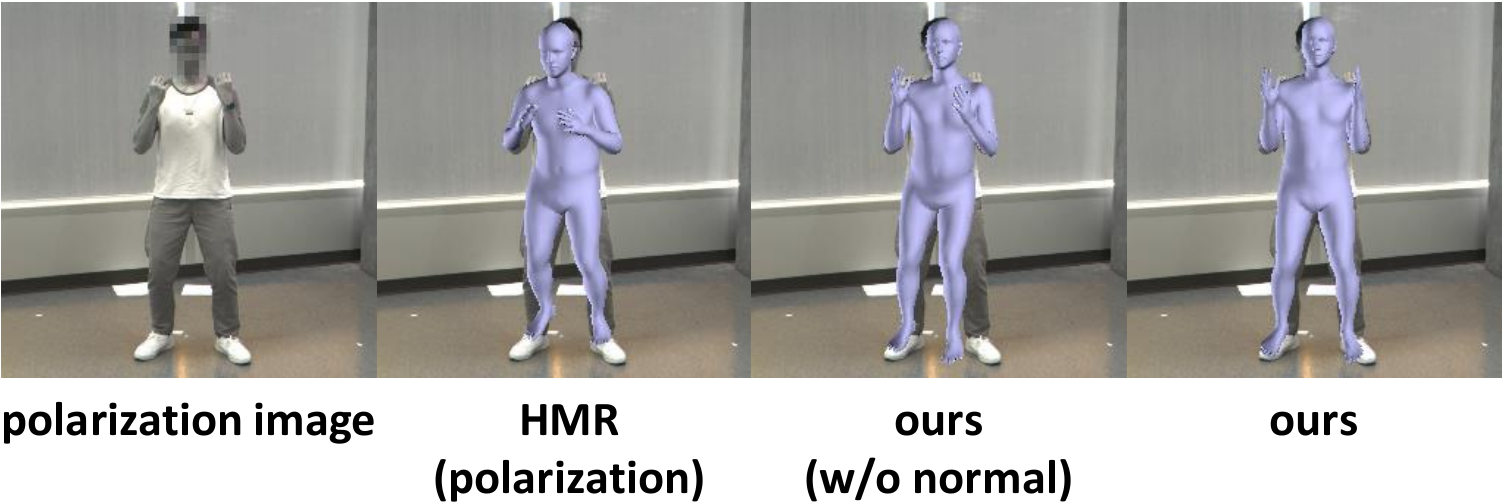}
    \caption{Exemplar shape estimation results. The first column is polarization images. HMR (polarization) means the HMR model is retrained on polarization images of our PHSPD dataset. Ours (w/o normal) means the model is trained without the normal map as a part of the input.}
    \label{fig:pose-estimation-demo}
\end{figure}

In pose estimation, it is of interest to inspect the effect of engaging surface normal maps in our SfP approach. %is an informative prior for better estimation of human poses and shapes. 
Besides our SfP approach, the competing methods consist of \emph{HMR}~\cite{kanazawa2018end} and a ablation variant of SfP, \emph{ours (w/o normal)}. The latter is obtained by engaging only the polarization image, without considering normal map estimation. Since \emph{HMR} is trained on single color images, it is re-trained using the first three channels of a polarization image. In addition to \emph{HMR} that works on color images, for fair comparison, \emph{HMR} is also re-trained on the polarization images of our PHSPD dataset, as \emph{HMR (polarization)}. From Table~\ref{tab:joint-estimation-polarization}, it is observed that our method produces the lowest MPJPE values of all competing methods; the results of \emph{ours (w/o normal)} is comparable to those of \emph{HMR (polarization)}. The quantitative results confirm that the polarization images is capable of producing accurate estimation of human poses. 
Moreover, the visual results in Fig.~\ref{fig:pose-estimation-demo} provide qualitative evidence that further performance gain is to be expected, when we have access to the normal maps. 
Similar observation is again obtained in Table~\ref{tab:joint-estimation-normal}, when quantitative examination is systematically carried out over w/ and w/o estimated normal map, on color and polarization images, in both datasets. Note the performance gain is particularly significant for polarization images, which may attribute to the rich geometric information encoded in the normal map representation. On color images, there is still noticeable improvement, also less significant. Our explanation is that the normal maps estimated from color images are not as reliable as those obtained from the polarization image counterparts.

\begin{table}[htb]
    \centering
    \begin{tabular}{p{90pt}p{60pt}p{60pt}p{60pt}}
    \toprule
        & \makecell[c]{SURREAL} & \multicolumn{2}{c}{PHSPD} \\
        \cmidrule{2-4}
        & \makecell[c]{GT-t} & \makecell[c]{GT-t} & \makecell[c]{Pred-t}  \\
    \midrule
        \makecell[c]{HMR \cite{kanazawa2018end}} & \makecell[c]{116.68/136.32} & \makecell[c]{82.96/91.46} & \makecell[c]{-}  \\ 
        \makecell[c]{HMR (polarization)} &  \makecell[c]{-} & \makecell[c]{77.57/88.74} & \makecell[c]{97.24/106.20}  \\
        \makecell[c]{ours (w/o normal)} &  \makecell[c]{83.43/94.00} & \makecell[c]{84.44/96.42} & \makecell[c]{93.38/104.48} \\
        \makecell[c]{ours} & \makecell[c]{\textbf{67.25/75.94}} & \makecell[c]{\textbf{66.32}/\textbf{74.46}} & \makecell[c]{\textbf{74.58}/\textbf{81.85}} \\
    \bottomrule
    \end{tabular}
    \caption{Quantitative evaluations using MPJPE evaluation metric on both SURREAL and PHSPD datasets. The unit of the error is millimeter. GT-t means the camera translation is known and Pred-t means the predicted camera translation is used to compute the joint error. We report the MPJPE results of 20/24 joints, which removes two hand and two foot joints following similar settings of previous work~\cite{kanazawa2018end,human36m}.}
    \label{tab:joint-estimation-polarization}
\end{table}

\begin{table}[htb]
    \centering
    \begin{tabular}{ccccc}
    \toprule
        & \multicolumn{2}{c}{SURREAL} & \multicolumn{2}{c}{PHSPD} \\
        \cmidrule{2-5}
        & ours (w/o normal) &\quad ours \quad & ours (w/o normal) & \quad ours \quad\\
    \midrule
        polarization image & 83.43/94.00 & 67.25/75.94 & 84.44/96.42 & 66.32/74.46   \\
        improvement & \multicolumn{2}{c}{\textbf{16.18/18.06}} & \multicolumn{2}{c}{\textbf{18.12/21.96}} \\
    \midrule
        color image & 88.53/100.32 & 80.70/91.51 & 85.67/80.34 & 77.72/70.07  \\
        improvement & \multicolumn{2}{c}{\textbf{7.82/8.81}} & \multicolumn{2}{c}{\textbf{7.95/10.27}} \\
    \bottomrule
    \end{tabular}
    \caption{Qualitative ablation study of our SfP approach (w/ vs. w/o the estimated surface normal). MPJPE is the evaluation metric with millimeter unit. Experiments are carried out on both color and polarization images of SURREAL and PHSPD datasets. %It demonstrates the effectiveness of having a fine surface normal map in human pose estimation.
    }
    \label{tab:joint-estimation-normal}
\end{table}

To evaluate the effectiveness of our approach on clothed human shape recovery, the state-of-the-art methods on human surface reconstruction from single color images are recruited. They are \emph{PIFu}~\cite{saito2019pifu}, \emph{Depth Human}~\cite{tang2019neural} and \emph{HMD}~\cite{zhu2019detailed}. Quantitative results are obtained in the PHSPD dataset by computing the 3D surface error of the predicted human mesh with respect to the ground-truth mesh. Scaled rigid ICP is performed before the evaluation so as to scale and transform the predicted mesh into the same coordinates as the ground-truth surface. The results are displayed in Table~\ref{tab:shape-estimation}. \emph{PIFu}~\cite{saito2019pifu} performs the worst, partly as it does not take human pose into consideration when predicting the implicit surface function inside a volume. The 3D surface error from \emph{HMD}~\cite{zhu2019detailed} and Depth \emph{Human}~\cite{tang2019neural} are relatively small; our SfP approach achieves the best performance, which is partly due to its exploitation of the estimated normal maps.
Note the comparison methods of \emph{PIFu}~\cite{saito2019pifu}, \emph{Depth Human}~\cite{tang2019neural} and \emph{HMD}~\cite{zhu2019detailed} only work with color images as input. 
In this experiment, for each of the polarization images used by the two SfP variants, namely \emph{our (w/o deform)} and \emph{ours}, the closet color image captured in the multi-camera setup of PHSPD is taken as its corresponding input to the three comparison methods.

\begin{table}[]
    \centering
    \begin{tabular}{p{90pt}p{35pt}p{60pt}p{35pt}p{68pt}p{35pt}}
    \toprule
         & \makecell[c]{PIFu\\\cite{saito2019pifu}} & \makecell[c]{Depth Human\\\cite{tang2019neural}} & \makecell[c]{HMD\\\cite{zhu2019detailed}} & \makecell[c]{ours\\(w/o deform)} & \makecell[c]{ours} \\
    \midrule
        3D surface error(mm) & \makecell[c]{73.13} & \makecell[c]{51.02} & \makecell[c]{51.71} & \makecell[c]{41.10} & \makecell[c]{\textbf{38.92}}\\
    \bottomrule
    \end{tabular}
    \caption{Quantitative evaluation of clothed human shape recovery performance methods in the PHSPD dataset. 
    %PIFu \cite{saito2019pifu}, Depth Human \cite{tang2019neural} and HMD \cite{zhu2019detailed} are the state-of-the-art methods on human surface reconstruction from single color images. 
    %on detailed human shape recovery from a single image and evaluate the 3D surface reconstruction error. The 3D error is computed as the mean distance from the ground-truth surface captured from three-view Kinects in our PHSPD dataset to the predicted mesh. The unit of the error is millimeter.}
    }
    \label{tab:shape-estimation}
\end{table}

%PIFu \cite{saito2019pifu} and Depth Human \cite{tang2019neural} are able to recover the human mesh with some surface details revealed, but the small details are still missing.

\begin{figure}[]
    \centering
    \includegraphics[width=\columnwidth]{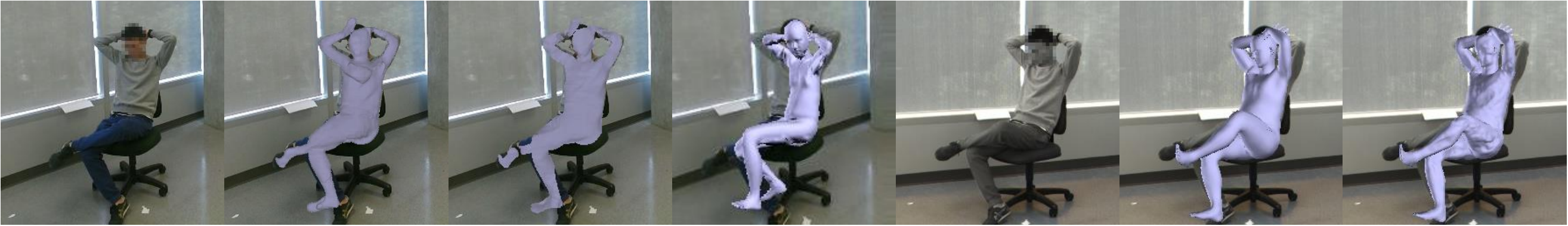}
    \includegraphics[width=\columnwidth]{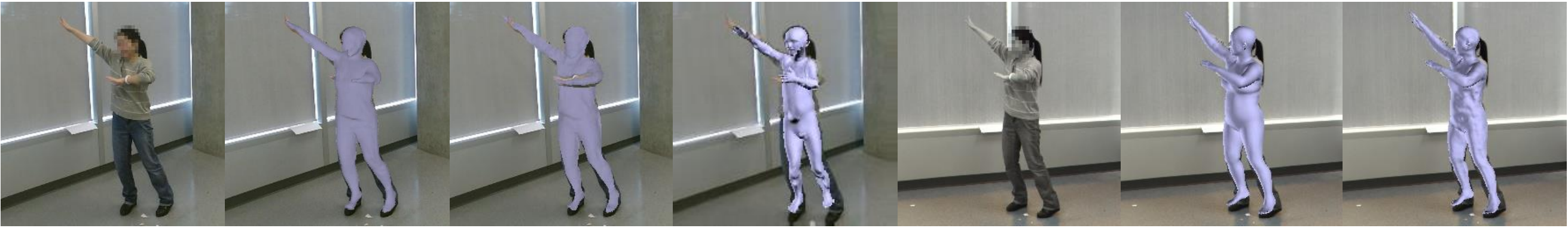}
    \includegraphics[width=\columnwidth]{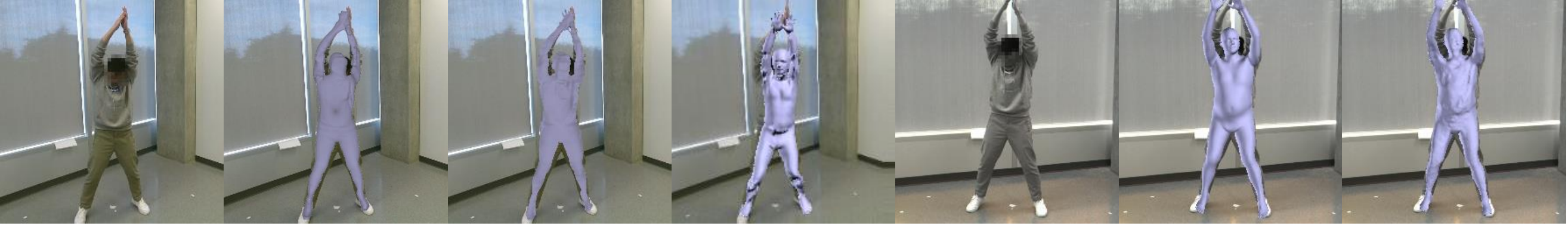}
    \includegraphics[width=\columnwidth]{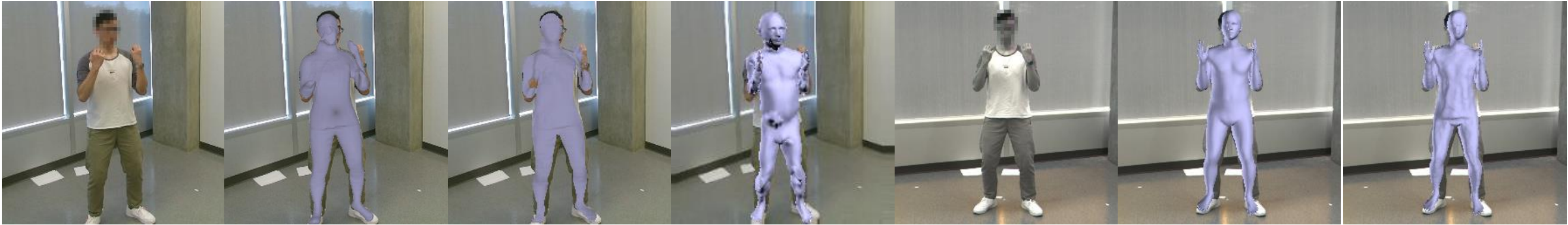}
    \includegraphics[width=\columnwidth]{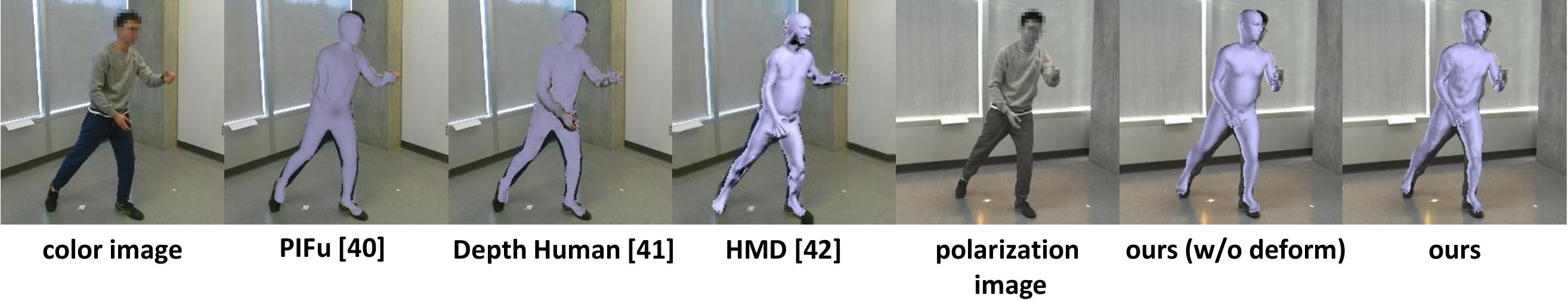}
    \caption{Exemplar estimation results of clothed body shapes. The first and fifth column are color images and polarization images, respectively. \emph{PIFu}~\cite{saito2019pifu}, \emph{Depth Human}~\cite{tang2019neural} and \emph{HMD}~\cite{zhu2019detailed} are the results based on color input images. \emph{Ours (w/o deformation)} and \emph{ours} are the results with the polarization image as the input.}
    \label{fig:detailed-shape-demo}
\end{figure}

Exemplar visual results are presented in Fig.~\ref{fig:detailed-shape-demo}, where the predicted body shapes are overlaid onto the input images. It is observed that the body shapes predicted by \emph{PIFu} and \emph{Depth Human} are generally well-aligned with the input image as they are actually predicting the implicit function value or depth value for each pixel of the foreground human shape. Meanwhile, it does not necessarily indicate accurate alignment of 3D surface mesh, as is evidenced in Table~\ref{tab:shape-estimation}. 
For \emph{PIFu} and \emph{Depth Human}, the exterior surfaces tend to be overly smooth. Besides, in \emph{Depth Human}, only a partial mesh with respect to the view in the input image is produced. \emph{HMD}, on the other hand, does not work well, as evidenced by the often error-prone surface details. This may be attributed to the less reliable shading representation, given the new environmental lighting and texture ambiguities existed in these color images. 
Our SfP approach is shown capable of producing reliable prediction of clothed body shapes, which again demonstrates the applicability of polarization imaging in shape estimation, as well as the benefit of engaging the surface normal maps in our approach.
%is able to obtain complete human meshes with real geometric surface details as we take advantage of the physical priors from the polarization image for surface reconstruction. 

% \begin{table}[htb]
%     \centering
%     \begin{tabular}{ccc}
%     \toprule
%          & polarization & color\\
%     \midrule
%         PIFu~\cite{saito2019pifu} & 71.65 & 73.13 \\
%         Deep Human~\cite{tang2019neural} & 59.48 & 51.02 \\
%         HMD~\cite{zhu2019detailed} &  49.94 & 51.71 \\
%         ours w/o deform & 41.10 & \textbf{50.23}\\
%         ours deform &  \textbf{38.92} & - \\
%     \bottomrule
%     \end{tabular}
%     \caption{Shape estimation. Note that our model does not require the prior knowledge of human body segmentation.}
%     \label{tab:shape-estimation}
% \end{table}

\begin{figure}[]
    \centering
    \includegraphics[width=\columnwidth]{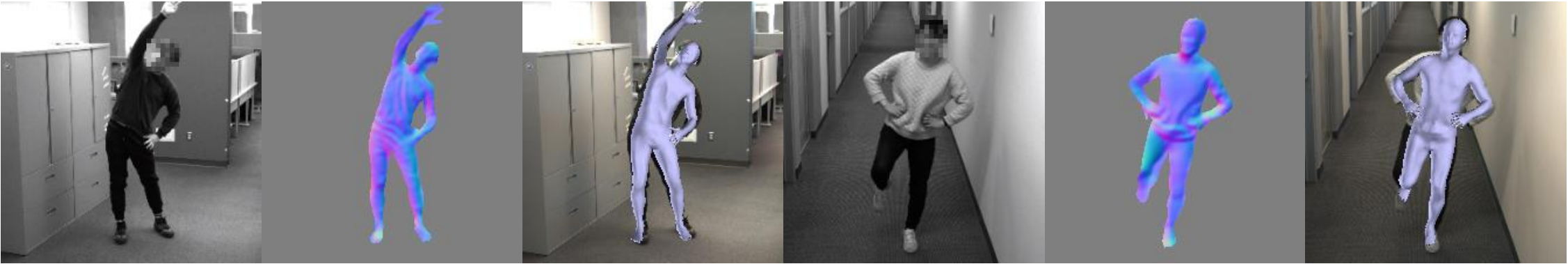}
    \caption{Exemplar estimation results of clothed body shapes, obtained on polarization images from novel test scenarios (new human subject and scene context).}
    \label{fig:extra-demo}
\end{figure}

Qualitative results presented in Fig.~\ref{fig:extra-demo} showcase the robust test performance in novel settings. Note the polarization images are intentionally acquired from unseen human subjects at new geo-locations, so the background scenes are very different from those in the training images. 

\section{Conclusion}
This paper tackles a new problem of estimating clothed human shapes from single 2D polarization images. 
Our work demonstrate the applicability of engaging polarization cameras as a promising alternative to the existing imaging sensors for human pose and shape estimation. 
Moreover, by exploiting the rich geometric details in the surface normal of the input polarization images, our SfP approach is capable of reconstructing clothed human body shapes of surface details. 
%We expect this could be a useful tool in many downstream applications. 
%For future work, we plan to thoroughly evaluate the performance of polarization, color, and depth imaging, over a broader range of image scenarios and human body poses.

\section*{Acknowledgement}
This work is supported by the NSERC Discovery Grants, %(No. RGPIN-2019-04575), 
and the University of Alberta-Huawei Joint Innovation Collaboration grants. 

\newpage
\bibliographystyle{splncs}
\bibliography{egbib}

\include{supplementary}
\end{document}

%% file: supplementary.tex
% \documentclass[runningheads]{llncs}
% \usepackage{graphicx}
% \usepackage{amsmath,amssymb} % define this before the line numbering.
% % \usepackage{ruler}
% \usepackage{booktabs}
% \usepackage{makecell}
% \usepackage{color}
% % \usepackage[width=122mm,left=12mm,paperwidth=146mm,height=193mm,top=12mm,paperheight=217mm]{geometry}
% \begin{document}

% \linenumbers
% \pagestyle{headings}
% \mainmatter
% \def\ECCV20SubNumber{2136}  % Insert your submission number here
\title{Supplementary Material}

\titlerunning{Supplementary Material}

% \author{Anonymous ECCV submission}
% \institute{Paper ID \ECCV20SubNumber}

\author{Shihao Zou\inst{1} \and
Xinxin Zuo\inst{1} \and
Yiming Qian\inst{2} \and
Sen Wang\inst{1} \and
Chi Xu\inst{3} \and
Minglun Gong\inst{4} \and
Li Cheng\inst{1}}
\authorrunning{S. Zou, et al.}

\institute{University of Alberta \and Simon Fraser University \and China University Of Geosciences \and University of Guelph\\
\email{\{szou2,xzuo,sen9,lcheng5\}@ualberta.ca,yimingq@sfu.ca,\\xuchi@cug.edu.cn,minglun@uoguelph.ca}}

\maketitle

\section{Implementation Details}
% architecture
The encoder-decoder model architecture is used in the stage of normal estimation. In the stage of shape and pose estimation, we use ResNet50 \cite{he2016deep} and the average-pooled output is directly regressed to an 85-dimension vector. The polarization image and the predicted normal map are concatenated as the input to estimate the SMPL shape parameters. ResNet50 is trained from scratch.

We synthesize a polarization image (polarizers of 0, 45, 90 and 135 degree) given the rendered depth and color image. In detail, from the depth image, we obtain the normal map and calculate the zenith and azimuth angle, and from the color image, we get the gray image and take it as the polarization image of 0 degree polarizer, denoted by $I(0)$. Assuming diffuse reflection of the human body surface, we can calculate the degree of polarization $\rho$ according to the equation,
\begin{equation}
    \label{eq:zenith-angle}
    \rho = \frac{(n-\frac{1}{n})^2\sin^2\theta}{2+2n^2-(n+\frac{1}{n})^2\sin^2\theta+4\cos\theta\sqrt{n^2-\sin^2\theta}},
\end{equation}
with the zenith angle and refractive index known. Then the upper and lower bound of the illumination intensity $I_{max}$ and $I_{min}$ can be solved in closed-form with the constraints,
\begin{equation}
    \label{eq:degree-of-polarization}
    \rho = \frac{I_{max} - I_{min}}{I_{max} + I_{min}},
\end{equation}
and
\begin{equation}
    \label{eq:polar}
    I(0)=\frac{I_{max} + I_{min}}{2} + \frac{I_{max} - I_{min}}{2} \cos (2\varphi)).
\end{equation}
Finally, we can use the equation, 
\begin{equation}
    \label{eq:polar}
    I(\phi_{pol})=\frac{I_{max} + I_{min}}{2} + \frac{I_{max} + I_{min}}{2} \cos (2(\phi_{pol}-\varphi)),
\end{equation}
to get the image for polarizer $\phi_{pol}$ of degree 45, 90 and 135. To make it close to the real-world applications, we add Gaussian noise with $\sigma=1/255$ to each pixel of the synthetic polarization image and then quantize the intensity value to 8 bits. Due to the fact that we only have geometric information for human body, the synthetic polarization images only have values on human body part.

We first train the normal estimation model for 20 epochs by setting $\lambda_c$ and $\lambda_n$ to be 2 and 1 respectively. The learning rate starts at $0.001$ and decays to $0.0001$ after 15 epochs. Then we train the shape estimation model for 30 epochs by setting $\lambda_{\beta}$, $\lambda_{\theta}$, $\lambda_{t}$ and $\lambda_{J}$ to be 0.2, 0.5, 100 and 3 respectively. The learning rate starts at $0.001$ and decays to $0.0001$ after 5 epochs. Adam optimizer \cite{kingma2014adam} is used to train our model.

To deform the SMPL based human mesh model towards the predicted normal map, first we integrate a depth map from the predicted normal map with the projected SMPL mesh as the coarse depth. In detail, we define a objective function as
\begin{equation}
    E(D) = \lambda_n E_\mathbf{n}(D) + \lambda_d E_d(D) + \lambda_s E_s(D),
\label{Eq:integrate}
\end{equation}
and we get the detailed depth via minimization of this function.

The first term, $E_\mathbf{n}(D)$, is used to enforce the predicted normal to be perpendicular to the tangents of the optimized depth surface,
\begin{equation}
     E_\mathbf{n}(D) = \sum_i T_x^i n_i + T_y^i n_i.
\end{equation}

The tangents $T_x$ and $T_y$ are defined as below,
\begin{equation}
     T_x = [\frac{1}{f_x}(\frac{\partial D}{\partial x}(x-p_x) +D ), \frac{1}{f_y}\frac{\partial D}{\partial x}(y-p_y),  \frac{\partial D}{\partial x}]^T,
\end{equation}

\begin{equation}
     T_y = [ \frac{1}{f_x}\frac{\partial D}{\partial x}(y-p_y), 
     \frac{1}{f_y}(\frac{\partial D}{\partial y}(y-p_y) +D ),
     \frac{\partial D}{\partial y}]^T.
\end{equation}

In the above function, $f_x$ and $f_y$ are the focal length and $p_x$ and $p_y$ are the camera center of the camera. 

For the second term $E_d(D)$, we set the boundary constraints so that the optimized depth to be close to the base depth $\hat{D}_i$,

\begin{equation}
     E_d(D) = \sum_i \Big( \big( (\frac{x_i-p_x}{f_x})^2 + (\frac{y_i-p_y}{f_y})^2 +1) (D_i - \hat{D}_i \big) \Big)^2.
\end{equation}

Finally we want to preserve smoothness for the integrated surface and add the smoothness constraints for neighboring pixels in the third term $E_s(D)$,

\begin{equation}
     E_s(D) = \sum_{i,j\in N} || D_i-D_j||^2.
\end{equation}

We find a linear least squares solution of the objective Eq.~(\ref{Eq:integrate}) and get the detailed depth map.

\section{More results}

\begin{figure}[]
    \centering
    \includegraphics[width=\columnwidth]{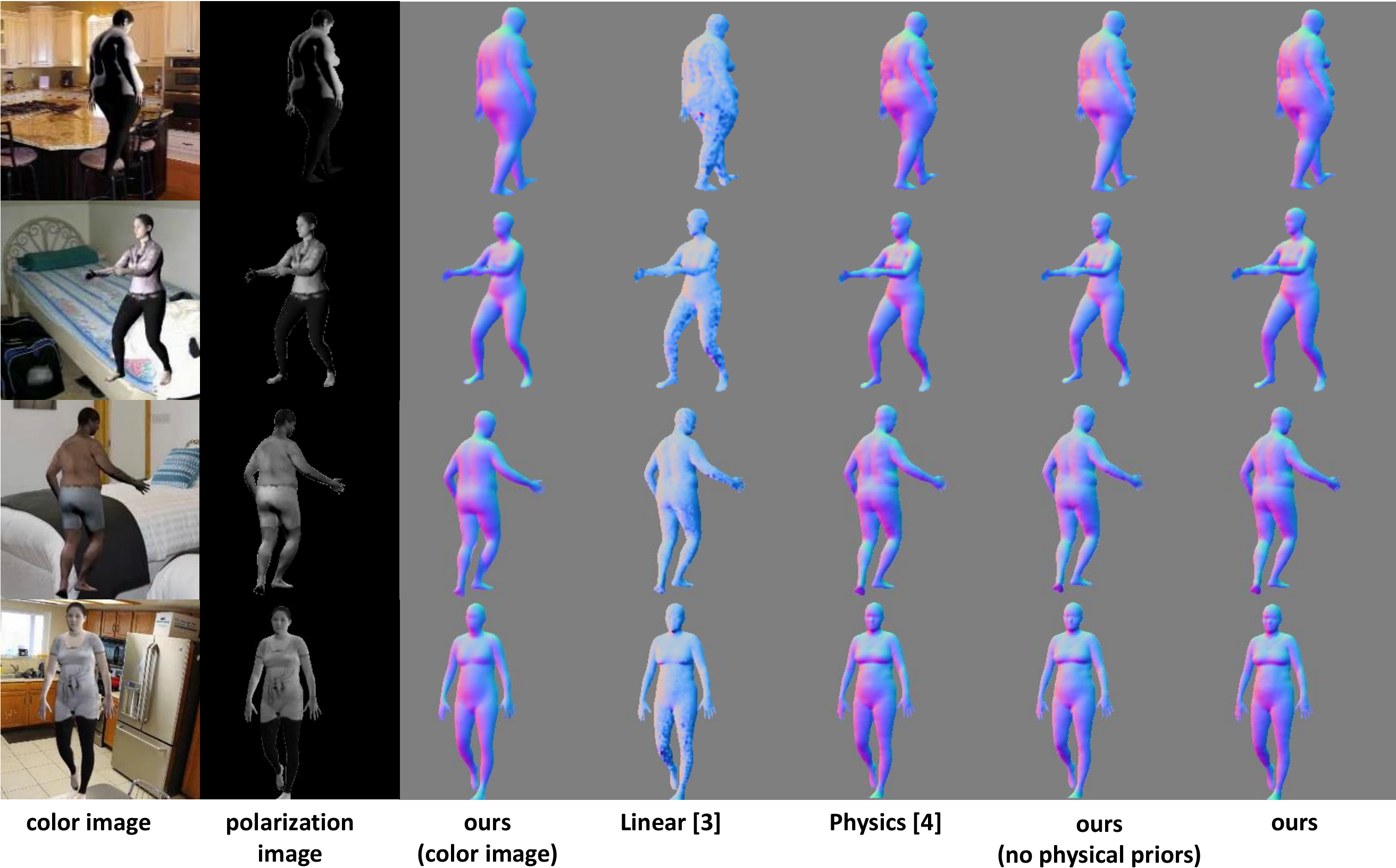}
    \caption{The figure shows the results of normal estimation on SURREAL dataset. The first column is color image for better visualization. The second column is the synthetic polarization image as the input to estimate normal map. The third column is the result from ours (color image). The fourth and fifth columns are the results from Linear \cite{smith2016linear} and Physics \cite{ba2019physics}. The sixth column is ours (no physical priors). Compared with ours (color image), we can see that the better surface normal is predicted by ours.}
    \label{fig:surreal-normal}
\end{figure}

\begin{figure}[]
    \centering
    \includegraphics[width=\columnwidth]{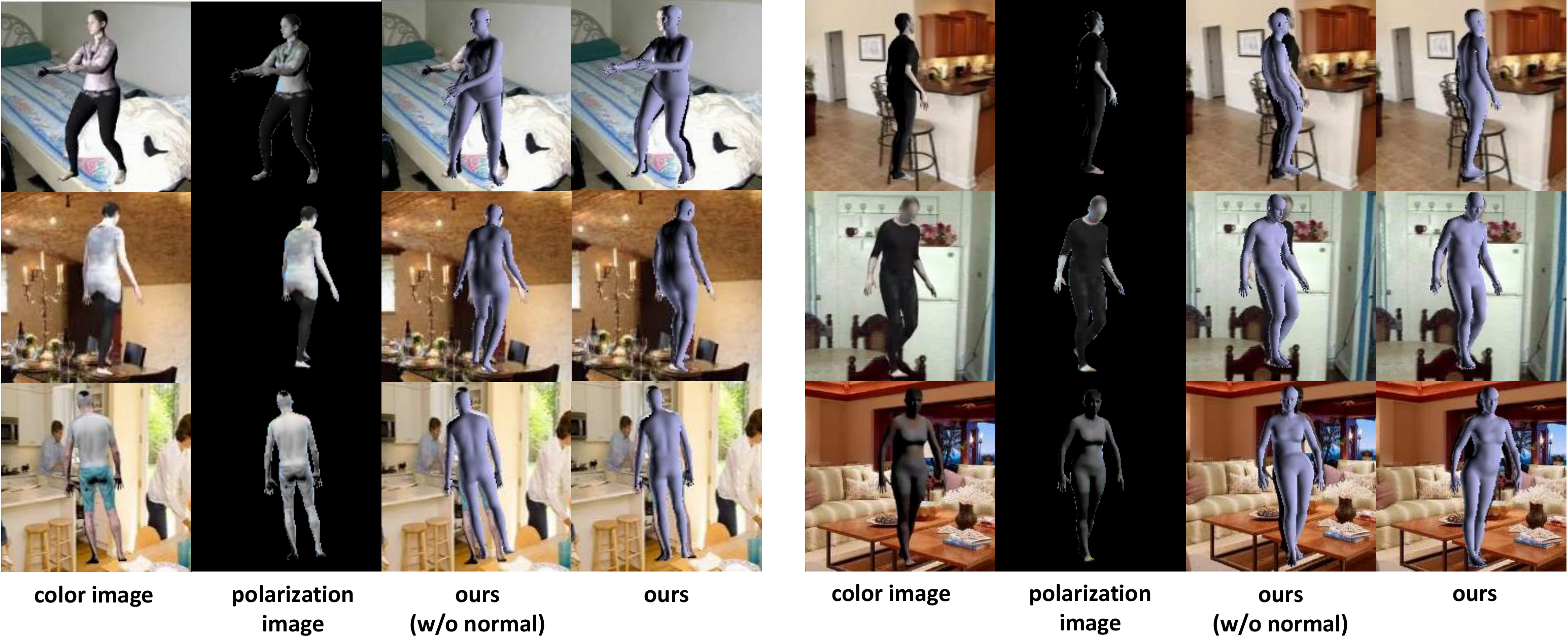}
    \caption{The figure shows the results of shape and pose estimation on SURREAL dataset. We show the color image in the first column for better visualization. The second column is polarization image, which is the input to estimate the shape. The third column is the result from ours (w/o normal). We can see that normal map is an informative priors to learn to predict better human shape from a polarization image.}
    \label{fig:surreal-pose}
\end{figure}

We show the results of normal estimation and shape estimation on SURREAL dataset in Fig.~\ref{fig:surreal-normal} and \ref{fig:surreal-pose} respectively. We display the corresponding color image in the first column for better demonstration. The results of Linear \cite{smith2016linear} and Physics \cite{ba2019physics} are also shown. The synthetic polarization images (displayed the first channel as a gray image) are shown in the second column, which are the input to estimate normal map and also the shape and pose. Due to the fact that we only have geometric information for human body, the synthetic polarization images only have values on human body part.

A video\footnote{The video, named \emph{detailed\_shape\_demo.avi}, is submitted in the supplementary material.} is presented to give a more comphrehensive view of our detailed shape. The video shows the predicted detailed human shape from different view angles. On the one hand, compared with PIFu \cite{saito2019pifu} and Depth Human \cite{tang2019neural}, we can see that our method is more robust to complex poses especially when we change the angle of view. On the other hand, compared with HMD \cite{zhu2019detailed}, our method can recover more reliable clothing details of human body.

\section{Polarization Human Shape and Pose Dataset (PHSPD)}
More details can be found on the site \footnote{\url{https://jimmyzou.github.io/publication/2020-PHSPDataset}}.
\subsection{Data Acquisition} 
\begin{figure}[]
    \centering
    \includegraphics[width=0.39\columnwidth]{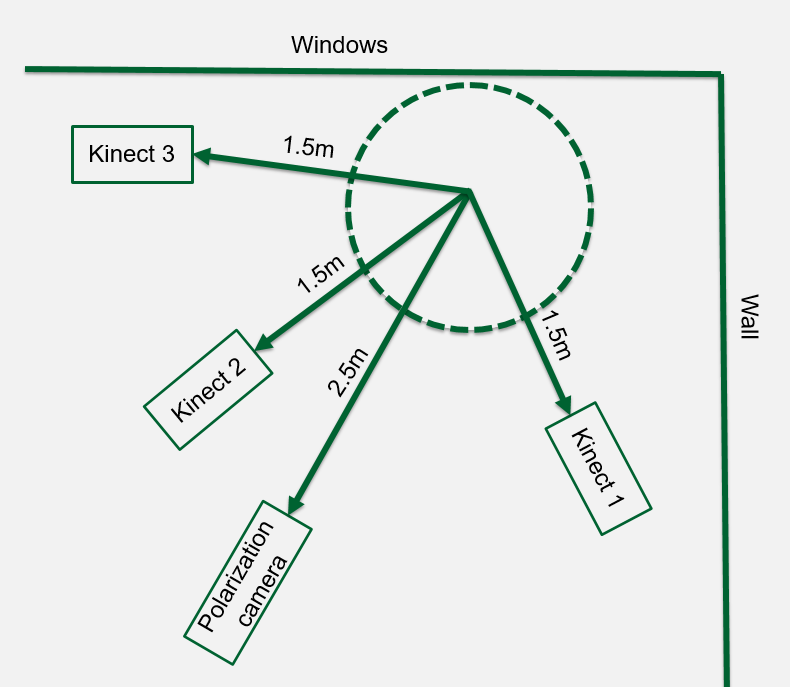}
    \includegraphics[width=0.59\columnwidth]{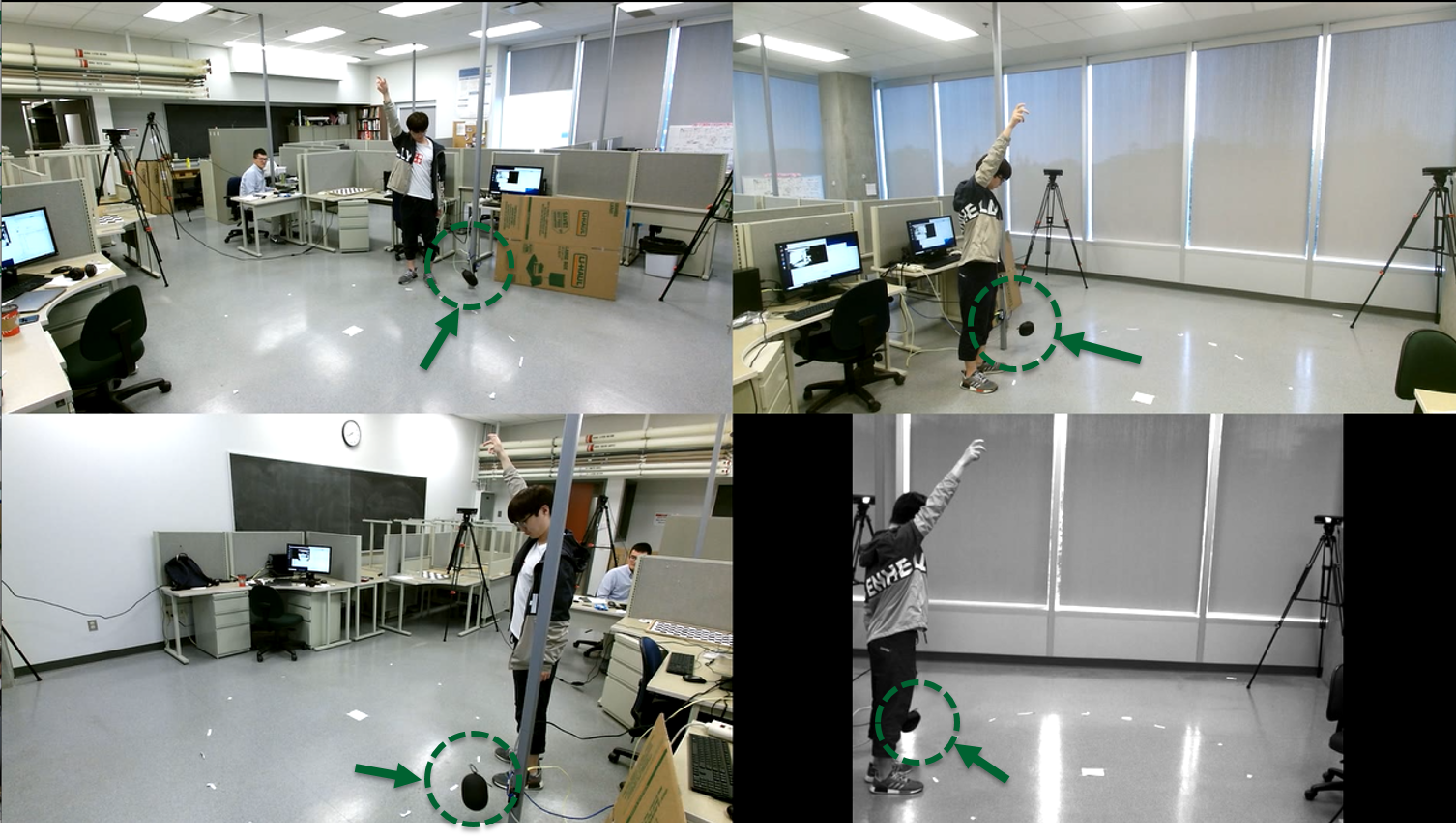}
    \caption{Left figure: the layout of our multi-camera system. Three Kinects are placed around a circle of motion area with one polarization camera. Right figure: the synchronization result of our multi-camera system. The same frame of the three-view color images and one-view polarization image are displayed. Note that the layout of our multi-camera system has been changed to the left figure, but other settings are the same.}
    \label{fig:camera-layout}
\end{figure}

Our acquisition system synchronizes four cameras, one polarization camera and three Kinects V2 in three different views (each Kinect v2 has a depth and a color camera). The layout is shown in Fig.~\ref{fig:camera-layout}. The other task is multi-camera synchronization. As one PC can only control one Kinect V2, we develop a soft synchronization method. Specifically, each camera was connected with a desktop (the desktop with the polarization camera is the master and the other three ones with three Kinects are clients). We use socket to send message to each desktop. After receiving certain message, each client will capture  the most recent frame from the Kinect into the desktop memory. At the same time, the master desktop sends a software trigger to the polarization camera to capture one frame into the  buffer. Fig.~\ref{fig:camera-layout} shows the synchronization performance of the system that we develop. We let a bag fall down and compare the position of the bag in the same frame from four views. We can find that the positions of the bag captured by four cameras are almost the same in terms of its distance to the ground.

\begin{table}[]
    \centering
    \begin{tabular}{|c|c|}
    \toprule
        group \# & actions \\
    \midrule
        1 &  warming-up, walking, running, jumping, drinking, lifting dumbbells\\
        2 &  sitting, eating, driving, reading, phoning, waiting\\
        3 &  presenting, boxing, posing, throwing, greeting, hugging, shaking hands\\
    \bottomrule
    \end{tabular}
    \caption{The table displays the actions in each group. Subjects are required to do each group of actions for four times, but the order of the actions each time is random.}
    \label{tab:dataset-action}
\end{table}
\begin{table}[]
    \centering
    \begin{tabular}{ccccc}
    \toprule
        \makecell[c]{\ subject\ \\\ \#\ } & \ gender\  & \makecell[c]{\ \ \# of original\ \ \\\  frames\ } & \makecell[c]{\ \ \# of annotated\ \ \\\  frames\ } & \makecell[c]{\ \ \# of discarded\ \ \\\  frames\ } \\
    \midrule
        % chenxiangye & fmale & 22561 & 22241 & 320 \\
        % cuihaibo & male & 24325 & 24186 & 139 \\
        % duanyuwei & male & 23918 & 23470 & 448 \\
        % guochuan & male & 24242 & 23906 & 336 \\
        % houpengyue & male & 24823 & 23430 & 1393 \\
        % lyuxingzheng & male & 24032 & 23523 & 509 \\
        % ranxiaomin & fmale & 22598 & 22362 & 236 \\
        % yangji & male & 23965 & 23459 & 506 \\
        % yanhe & male & 24712 & 24556 & 156 \\
        % zhouming & fmale & 24040 & 23581 & 459 \\
        % zoushihao & male & 24303 & 23795 & 508 \\
        % zouting & male & 24355 & 23603 & 752 \\
        1 & female & 22561 & 22241 & 320 (1.4\%) \\
        2 & male & 24325 & 24186 & 139 (0.5\%) \\
        3 & male & 23918 & 23470 & 448 (1.8\%)\\
        4 & male & 24242 & 23906 & 336 (1.4\%)\\
        5 & male & 24823 & 23430 & 1393 (5.6\%)\\
        6 & male & 24032 & 23523 & 509 (2.1\%)\\
        7 & female & 22598 & 22362 & 236 (1.0\%)\\
        8 & male & 23965 & 23459 & 506 (2.1\%)\\
        9 & male & 24712 & 24556 & 156 (0.6\%)\\
        10 & female & 24040 & 23581 & 459 (1.9\%) \\
        11 & male & 24303 & 23795 & 508 (2.1\%)\\
        12 & male & 24355 & 23603 & 752 (3.1\%)\\
        \midrule
        total & - & 287874 & 282112 & 5762 (2.0\%)\\
    \bottomrule
    \end{tabular}
    \caption{The table shows the detail number of frames for each subject and also the number of frames that have SMPL shape and 3D joint annotations.}
    \label{tab:dataset-details}
\end{table}
Our dataset has 12 subjects, 9 male and 3 female subjects. Each subject is required to do 3 different groups of actions (18 different actions in total) for 4 times plus one free-style group. Details are shown in Tab.~\ref{tab:dataset-action}. So each subject has 13 short videos and the total number of frames for each subject is around 22K. Overall, our dataset has 287K frames with each frame including one polarization image, three color and three depth images. Quantitative details of our dataset are shown in Tab.~\ref{tab:dataset-details}

\subsection{Annotation Process} 
The reason that we use multi-camera system to acquire image data is that multi-camera system provides much more information than a single-camera system. So the annotation of SMPL human shape and 3D joint position is more reliable using information of three-view Kinects v2. 

After camera calibration and plane segmentation of depth images, now we have a point cloud of human fused from three-view depth image and noisy 3D joint position by Kinect SDK at hand. The annotation SMPL human shape and 3D joint position has three main steps. First step is to filter out accurate 3D joint position by three Kinects in three views. For each view, we get the 2D joint estimation by OpenPose \cite{openpose} and also the 2D Kinect joint by projecting the noisy Kinect 3D joint to the color image. 

Then we compare the 2D distance between these two estimated joints. If the distance is larger than 50 pixel distance, we regard the joint estimated by the Kinect as incorrect one. As we have the joint estimation from three views, we simply average the correct joint position of three views and consider it as the initial guess of the position of the joint. If none of the three-view estimated joint is correct, we consider it as a missing joint. In this way, we get the initial guess of 3D joint positions for each frame and we discard the frame with more than 2 joints missing (14 in total). The next step is similar to \cite{bogo2016keep}, but instead of fitting to the 2D joints which have inherent depth ambiguity,  we fit SMPL model to the initial guess of 3D joints. 

Furthermore, as we have the point cloud of a human from three-view depth cameras, our final step is to further iteratively optimize SMPL parameters by minimizing the distance between vertices of SMPL shape to their nearest point. Finally, we have the annotated SMPL shape parameters and 3D joint positions. 

Besides, we render the boundary of SMPL shape on the image to get the mask of background, and calculate the target normal using three depth images based on \cite{qi2018geonet}. Although the target normal is noisy, our experiment result shows our model can still learn to predict good and smooth normal maps.

\begin{figure}[]
    \centering
    \includegraphics[width=0.8\columnwidth]{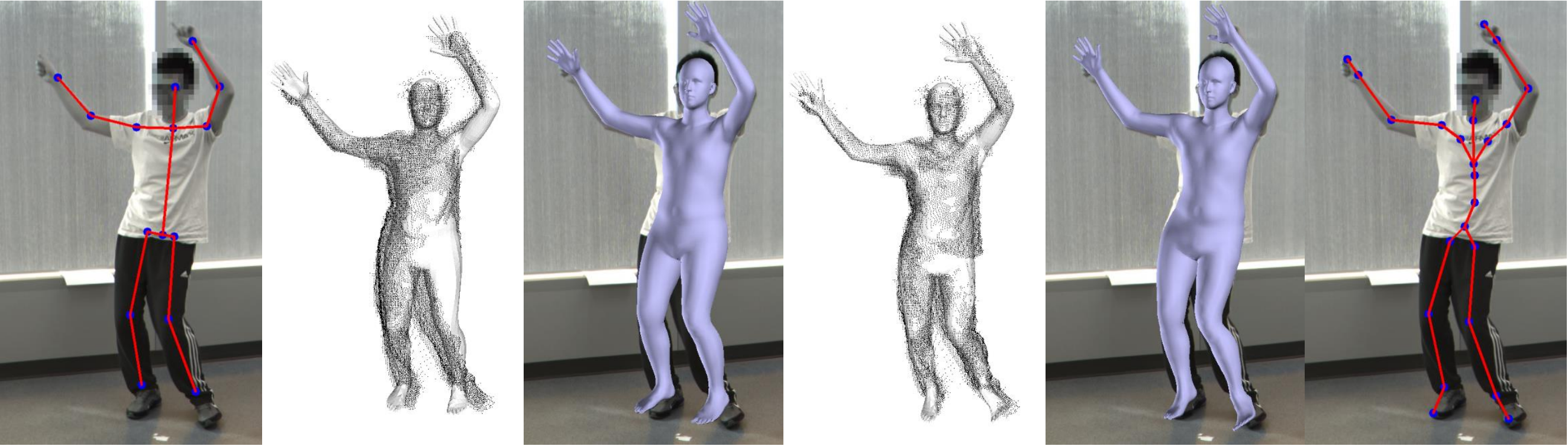}
    \includegraphics[width=0.8\columnwidth]{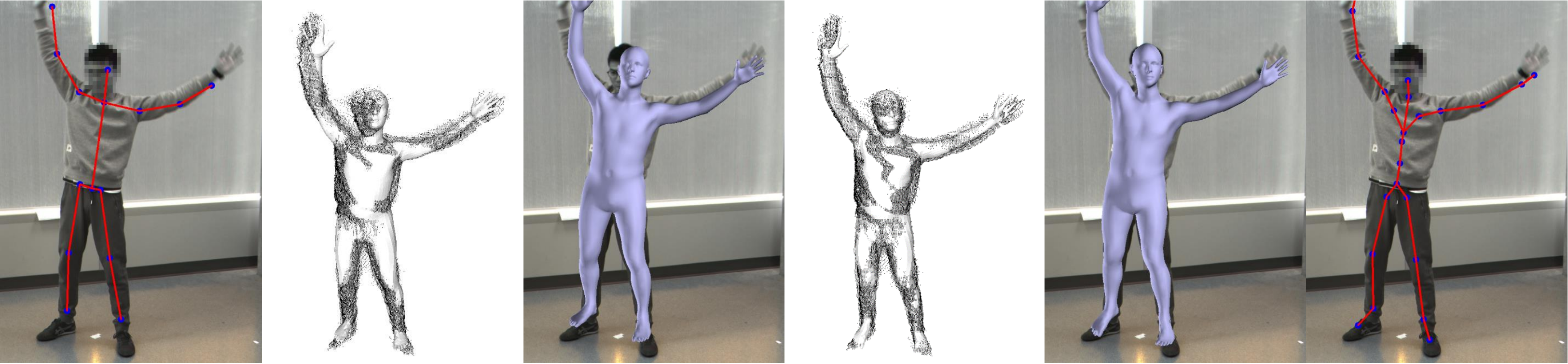}
    \includegraphics[width=0.8\columnwidth]{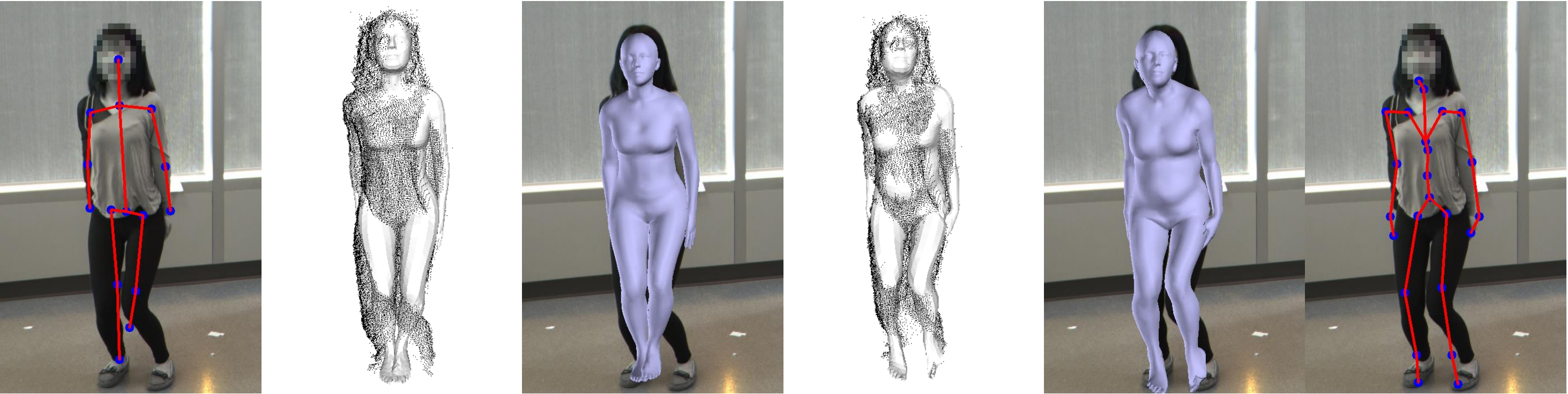}
    \includegraphics[width=0.8\columnwidth]{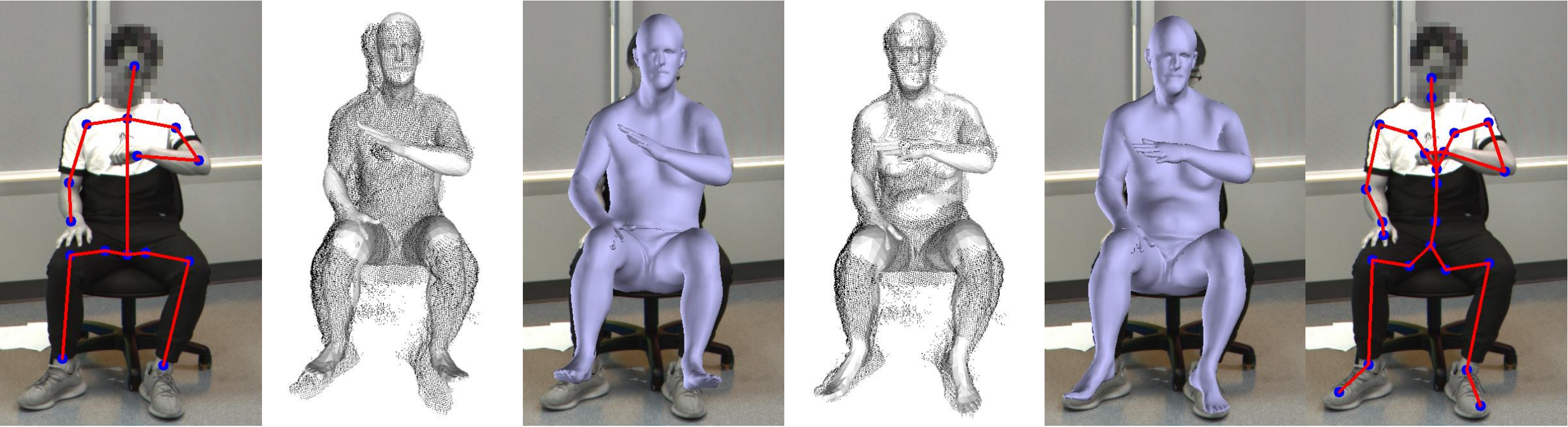}
    \includegraphics[width=0.8\columnwidth]{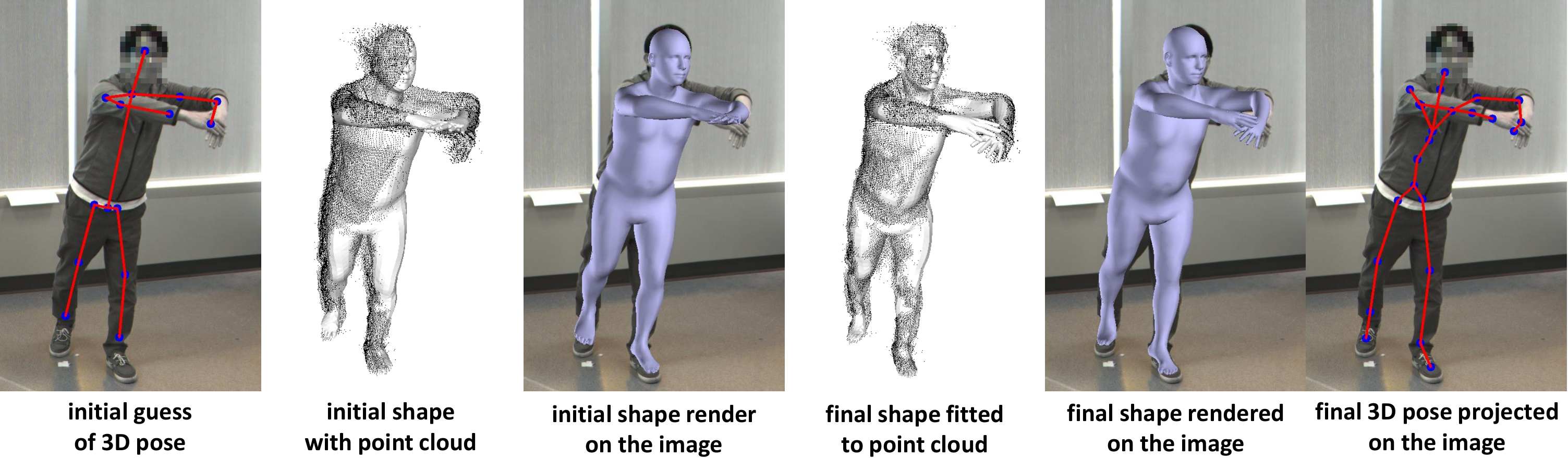}
    \caption{The figure shows our annotation process. The first column shows the initial guess of 3D pose, which is projected on the polarization image. After fitting the SMPL shape to the initial pose, we show the initial shape with the point cloud of human mesh (black points) in the second column and the rendered shape on the image in the third column. The fourth and fifth columns show the annotated shape after fitting to the point cloud of human mesh. The sixth column shows the corresponding annotated 3D pose.}
    \label{fig:annoataion-demo}
\end{figure}

\begin{figure}[]
    \centering
    \includegraphics[width=\columnwidth]{fig/supplement/fig2_1.pdf}
    \includegraphics[width=\columnwidth]{fig/supplement/fig2_2.pdf}
    \includegraphics[width=\columnwidth]{fig/supplement/fig2_3.pdf}
    \includegraphics[width=\columnwidth]{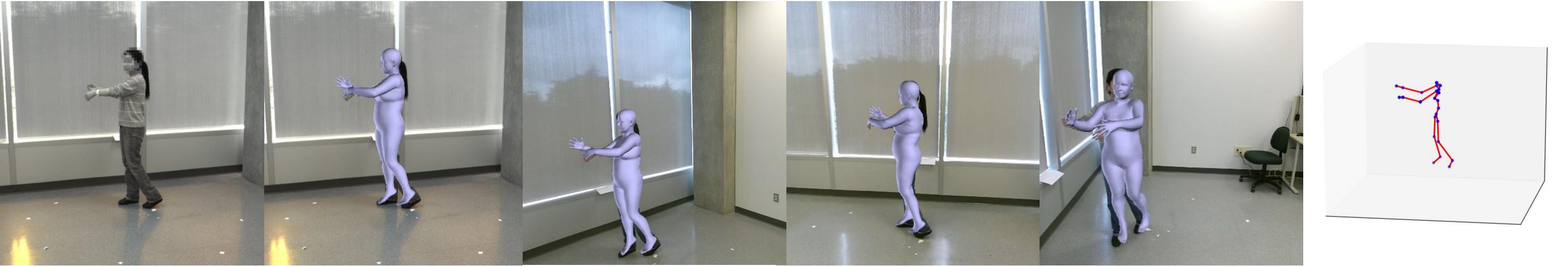}
    \includegraphics[width=\columnwidth]{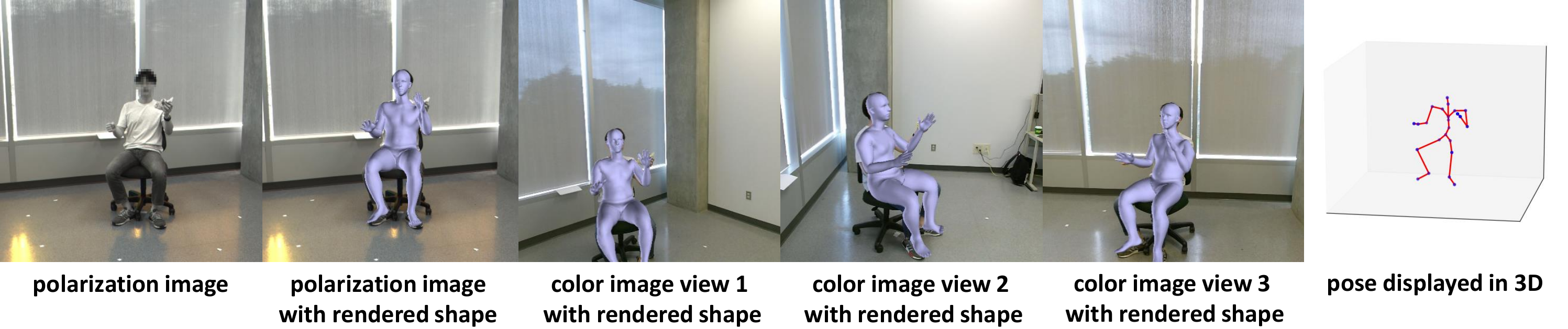}
    \caption{The figure shows our annotated shapes and poses. The first column is the polarization image for reference. The second to the fifth columns show the annotated shape rendered on the polarization image and three-view color images. The sixth column shows the annotated pose in 3D space.}
    \label{fig:multiview-demo}
\end{figure}

The annotation process is shown in Fig.~\ref{fig:annoataion-demo}. Starting from the initial guess of 3D pose, we fit SMPL shape to the initial 3D pose and further fit to the point cloud of human mesh. Finally, we get annotated human shape and pose. Besides, we also show our annotated shape on multi-view images (one polarization image and three-view color image) and the human pose in 3D coordinate space in Fig~\ref{fig:multiview-demo}. We also have a video \footnote{The video, named \emph{annotation\_demo.avi}, is submitted in the supplementary material.} to show our annotation results.

% \newpage
% \bibliographystyle{splncs}
% \bibliography{egbib}
% \end{document}